\documentclass[10pt,twocolumn,letterpaper]{article}

\usepackage[pagenumbers]{iccv} %

\usepackage{microtype}
\usepackage{multirow} 
\usepackage{tcolorbox}
\usepackage{bm}
\usepackage[misc]{ifsym}
\usepackage{xcolor}
\usepackage{fancyvrb}
\usepackage{makecell}
\usepackage{adjustbox}
\usepackage{fontawesome} %
\usepackage{xcolor}
\usepackage{colortbl}

\definecolor{gold}{rgb}{1, 0.75, 0.75} %
\definecolor{silver}{rgb}{1, 0.875, 0.75} %
\definecolor{bronze}{rgb}{1, 1, 0.75} %

\newcommand{\tgold}[1]{\colorbox{gold}{{#1}}}
\newcommand{\tsilver}[1]{\colorbox{silver}{{#1}}}
\newcommand{\tbronze}[1]{\colorbox{bronze}{{#1}}}

\newcommand{\gold}[1]{\cellcolor{red!15}{#1}}
\newcommand{\silver}[1]{\cellcolor{orange!15}{#1}}
\newcommand{\bronze}[1]{\cellcolor{yellow!15}{#1}}

\newcommand{\method}{{Exploring The Visual Feature Space for Multimodal Neural Decoding}}

\newcommand{\condparagraph}[1]{\vspace{0.25em}\noindent\textbf{#1}\enspace}

\definecolor{iccvblue}{rgb}{0.21,0.49,0.74}
\usepackage[pagebackref,breaklinks,colorlinks,allcolors=iccvblue]{hyperref}

\usepackage[capitalize]{cleveref}
\crefname{section}{Sec.}{Secs.}
\Crefname{section}{Section}{Sections}
\Crefname{table}{Table}{Tables}
\crefname{table}{Tab.}{Tabs.}
\Crefname{figure}{Figure}{Figures}
\crefname{figure}{Fig.}{Figs.}

\def\confName{ICCV}
\def\confYear{2025}

\title{\method}

\author{
Weihao Xia $^{\textrm{\Letter}}$ \quad Cengiz Oztireli \\[2mm]
 University of Cambridge \vspace{3pt}\\
\hypersetup{urlcolor=magenta}
{\small \url{https://weihaox.github.io/VINDEX}}
\vspace{-1em}
}

\begin{document}
\maketitle

\begin{abstract}
The intrication of brain signals drives research that leverages multimodal AI to align brain modalities with visual and textual data for explainable descriptions. However, most existing studies are limited to coarse interpretations, lacking essential details on object descriptions, locations, attributes, and their relationships. This leads to imprecise and ambiguous reconstructions when using such cues for visual decoding. To address this, we analyze different choices of vision feature spaces from pre-trained visual components within Multimodal Large Language Models (MLLMs) and introduce a zero-shot multimodal brain decoding method that interacts with these models to decode across multiple levels of granularities. %
To assess a model's ability to decode fine details from brain signals, we propose the Multi-Granularity Brain Detail Understanding Benchmark (MG-BrainDub). This benchmark includes two key tasks: detailed descriptions and salient question-answering, with metrics highlighting key visual elements like objects, attributes, and relationships. Our approach enhances neural decoding precision and supports more accurate neuro-decoding applications.
Code is available at \protect{\small \url{https://github.com/weihaox/VINDEX}}.
\end{abstract}

\vspace{-.3em}
\section{Introduction}
\vspace{-.2em}
\label{sec:intro}

Human brain signals are inherently complex and not directly interpretable, posing a significant challenge for decoding brain activity. To address this, recent studies have focused on translating brain signals associated visual stimuli into multimodal explanations, such as textual descriptions~\cite{scotti2024mindeye2,han2024onellm}, depth information~\cite{xia2024dream}, bounding boxes~\cite{xia2024umbrae}, or direct image reconstruction~\cite{scotti2024mindeye2,ozcelik2023brain,scotti2023reconstructing}. Although these efforts have made notable strides, achieving a fine-grained understanding of brain modalities remains an unresolved challenge.

One line of research~\cite{xia2024umbrae,takagi2023improving,scotti2024mindeye2} has attempted to translate brain activations from visual stimuli into textual captions. However, existing methods typically generate one-sentence descriptions with limited details, failing to capture the full spectrum of visual perception.
The most straightforward approach~\cite{shen2024neuro} is to use off-the-shelf MLLMs~\cite{liu2023visual,li2022blip} to generate detailed descriptions from paired images for each brain recording, then train a model to decode these pseudo-annotations from brain signals. However, this practice is computationally inefficient and may introduce bias or errors, making it impractical for generalizable brain decoding.
The other key issue is the evaluation of brain-decoded detailed captions, which can be hundreds of words long, rendering conventional metrics unsuitable. Specifically, traditional captioning metrics~\cite{papineni2002bleu,vedantam2015cider,banerjee2005meteor}
which mostly rely on $n$-gram matching between candidate and reference captions, are highly sensitive to stylistic variations, leading to inconsistent evaluations. Model-based metrics either use outdated text encoders~\cite{anderson2016spice}
or struggle with long text due to token length limitations~\cite{radford2021learning,hessel2021clipscore}.
While concurrent studies~\cite{shen2024neuro} attempt to generate detailed captions for visual decoding, they lack reliable evaluations of caption quality and accuracy.

In this paper, we address the two critical issues. For method, 
we investigate the exploitation of various feature spaces from the visual components within MLLMs to enable fine-grained multimodal explanations from neural activations.
We present VINDEX (VIsual experts for multimodal Neural DEcoding eXploration), which leverages diverse feature spaces for multimodal neural decoding. We further introduce a denoiser-based alignment strategy that efficiently maps brain tokens to image tokens, allowing for integration into pre-trained MLLMs. This approach facilitates decoding across multiple levels of granularities without the need for additional textual or spatial annotations.
For evaluation, we propose the Multi-Granularity Brain Detail Understanding Benchmark (MG-BrainDub) to discern a model's ability to decode fine-grained details from brain signals.  MG-BrainDub consists of two key tasks: detailed descriptions and salient question-answering, with metrics focusing on key visual elements, including objects, attributes, and relationships, derived from both candidate and ground truth captions. Our contributions are summarized as follows:
\begin{itemize}
    \item Through systematic analysis, we exploit representative feature spaces of visual components within MLLMs for brain signal alignment and multimodal brain interaction.  
    \item We introduce a zero-shot multimodal brain decoding method for multi-granular decoding without the need for additional textual or spatial annotations during training. %
    \item We construct a benchmark with metrics for fine-grained brain interpretation, evaluating a model's ability to decode detailed descriptions and perform complex reasoning.
\end{itemize}

\section{Related Work}
\label{sec:related_work}

\condparagraph{Multimodal Brain Perception.}
Recent advancements in decoding multimodal cues from brain signals, facilitated by MLLMs, have achieved unprecedented performance \cite{ozcelik2023brain,takagi2023improving,xia2024dream,scotti2023reconstructing}. These methods typically map brain responses, captured through functional magnetic resonance imaging (fMRI), to more commonly used modalities that can be processed by pre-trained models for subsequent information extraction \cite{karras2020analyzing,rombach2022high,xu2022versatile}.
For instance, Takagi and Nishimoto~\cite{takagi2023improving} connect fMRI data with CLIP text embeddings and the latent space of Stable Diffusion (SD) \cite{rombach2022high} using ridge regression to reconstruct visual stimuli. Xia~\etal~\cite{xia2024dream} extract semantic, depth, and color features from fMRI data to reconstruct images using a depth- and color-conditioned SD model.
Several methods~\cite{han2024onellm,takagi2023improving,xia2024dream,scotti2024mindeye2} aim to create readable multimodal explanations of visual stimuli, such as textual description or depth estimation, but often rely on corresponding external supervision to guide the multimodal training.
In contrast, UMBRAE~\cite{xia2024umbrae} has recently demonstrated the ability to directly decode textual descriptions and bounding boxes from fMRI data using only image-fMRI pairs during training, without requiring additional annotations.
This method aligns image features with visual encoders and integrates them into MLLMs. Building on this foundation, our proposed approach aims to  further elucidate feature space representations, extracting more granular information while introducing metrics for evaluation.

\condparagraph{Multimodal Large Language Models.} 
Expanding Large Language Models (LLMs) to incorporate additional modalities, such as images, has recently attracted significant interest~\cite{liu2023visual,dosovitskiy2020image,yao2025dense,tong2024eyes}.
These models typically comprise three components: a frozen image encoder, a trainable connector, and a frozen or finetuned LLM. The connector bridges the gap from the image feature space to the LLM token space, which can be implemented as a linear layer~\cite{chen2023shikra}, a multilayer perceptron (MLP)~\cite{liu2023visual}, or a lightweight transformer~\cite{jaegle2021perceiver}.
Recent studies have shown that using pre-trained CLIP~\cite{radford2021learning} as the visual component can lead to compromised performance in visual tasks, particularly in grounding capabilities~\cite{chen2023shikra,tong2024eyes}. To overcome this limitation, researchers have explored ways to more effectively utilize the visual component to enhance perception across diverse scenarios. This includes employing a mixture of vision  encoders~\cite{shen2025mome,zong2024mova,tong2024eyes,kar2024brave} pre-trained on various tasks as expert models and investigating the feature spaces~\cite{yao2025dense,cai2024matryoshka} within CLIP. Drawing inspiration from these efforts, we exploit brain alignment with feature spaces in MLLM's visual component for multi-granular representation to facilitate zero-shot multimodal brain decoding.

\condparagraph{Multimodal-Brain Alignment.} 
The prevailing practice for brain alignment is to map neural modalities into a shared multimodal space~\cite{scotti2023reconstructing,ozcelik2023brain,xia2024umbrae,han2024onellm}, which can be categorized into generative alignment and discriminative alignment. 
The first category, exemplified by OneLLM~\cite{han2024onellm}, employs generative training to learn multimodal alignment, connecting multimodal inputs, including brain signals, with an LLM.
Methods in the second category align brain signals within a pre-trained embedding space, such as CLIP~\cite{radford2021learning}, using techniques like linear regression~\cite{takagi2023improving,ozcelik2023brain}, contrastive learning~\cite{xia2024dream}, diffusion priors~\cite{scotti2023reconstructing}, or feature reconstruction~\cite{xia2024umbrae}.
MEVOX~\cite{xia2025mevox} ensembles multiple task-specific experts to capture distinct aspects of brain perception. The brain encoder learns to align with omni-contextual explanations.
Most of these alignment strategies require per-subject training or subject-specific annotations, which can lead to scalability issues in practice. While UMBRAE~\cite{xia2024umbrae} performs zero-shot multimodal brain decoding, its regression-based alignment is deterministic. We analyze the limitations of this strategy and propose incorporating a novel denoising optimization objective to enhance the brain-feature alignment.

\begin{figure*}[t!]
    \centering
    \includegraphics[width=0.90\linewidth]{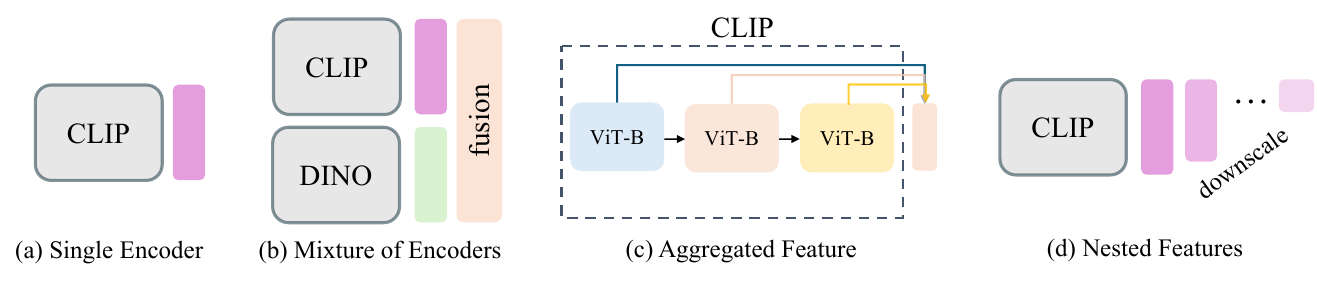}
    \vspace{-2mm}
    \caption{\textbf{Feature Spaces.} We aim to capture multi-granular representations for zero-shot multimodal brain decoding by aligning brain signals across different feature spaces in MLLM’s visual component. The explored representative feature spaces include: (a) Single Encoder, which utilizes a single encoder 
    to extract selected features; (b) Mixture of Encoders, using hybrid featured from different vision experts specialized in task-specific domains; (c) Aggregated Feature, fusing features from different layers within the same image encoder; and (d) Nested Features, downscaling visual representations in a hierarchical coarse-to-fine structure with different perceptual granularities.}
    \label{fig:feature_space}
\end{figure*}

\section{Method}
\label{sec:method}

Current MLLMs~\cite{liu2023visual,chen2023shikra} comprise three components: a pre-trained vision encoder, a pre-trained LLM, and a connector that bridges vision and language models. Given significant discrepancies in data scale and challenges associated with acquiring large-scale brain datasets, we argue that aligning brain signals with pre-trained visual components, rather than training from scratch~\cite{shen2024neuro,han2024onellm}, is a more effective and efficient approach. This not only alleviates the need to construct extensive text for training MLLMs but also enables zero-shot capability in decoding multimodal cues, allowing for versatile usage across different MLLMs. Thus, we present the rationale behind our method design in three key aspects: (1) the selection of vision feature spaces (\cref{subsec:feature_space}), (2) the alignment strategy to connect activations with pre-trained MLLMs (\cref{subsec:alignment_strategy}), and (3) the development of a benchmark necessary for fine-grained brain perception (\cref{sec:benchmark}).

\subsection{Exploring Feature Spaces} %
\label{subsec:feature_space}

We explore four distinct feature spaces, drawing inspiration from prior MLLM research that leverage different feature representations. By aligning brain signals across these feature spaces, we aim to capture multi-granular representations in MLLMs, facilitating zero-shot multimodal brain decoding. The four feature space choices are illustrated in~\cref{fig:feature_space}.

\condparagraph{Single Encoder.} This represents using features from a single pre-trained vision encoder (\eg~CLIP~\cite{radford2021learning}) for brain alignment, which is the most commonly used in MMLMs.

\condparagraph{Mixture of Encoders.} %
Encoders trained on diverse tasks allow us to explore the distinct advantages of different vision experts. These include contrastive image-text learning CLIP~\cite{radford2021learning}, self-supervised learning DINO~\cite{oquab2023dinov2}, and object-centric models pre-trained for various tasks such as segmentation (SAM~\cite{kirillov2023segment}), detection (EVA~\cite{fang2023eva}), and recognition (Pix2Struct~\cite{lee2023pix2struct}). These task-specific vision encoders allow MLLMs to achieve optimal performance within respective pre-training domains~\cite{kar2024brave,shi2024eagle}. Aligning brain signals with features from these models allows for the capture of distinct facets of visual perception. Here, we explore the integrated features from CLIP~\cite{radford2021learning} and DINO~\cite{oquab2023dinov2} encoders for simplicity while demonstrating the core concept of aligning brain signals with a mixture of vision experts. CLIP effectively encodes semantic information, while DINO excels in capturing spatial details such as object location and boundaries.

\condparagraph{Aggregated Feature.} %
Existing MLLMs typically use features from a selected layer in the visual component as the visual representation input for LLMs. Features from different levels—such as shallow, middle, and deep layers—are recognized for capturing different characteristics of the image~\cite{yao2025dense}. Therefore, by encapsulating information from multiple layers, we can enrich the visual input for the LLM.
Specifically, we partition features from selected layers into several groups. Features within each group from adjacent layers are integrated along the channel dimension to reduce redundancy and mitigate high dimensionality. The features from each group are then concatenated with the features from the final layer before being fed into the connector. 

\condparagraph{Nested Features.} %
MLLMs typically project feature images from the vision encoder into a fixed-length set of visual tokens~\cite{liu2023visual}. Longer token sequences generally yield better results but require more resources. Recent studies have found that the optimal token length is task-specific, laying the foundation for visual token pruning and merging to dynamically adjust the sequence length~\cite{bolya2022token,kim2024token}. Therefore, we downscale selected features using a set of downsampling factors and produce nested features. This way offers a flexible and controllable way to encode visual content as nested sets of visual tokens with a hierarchical structure, from coarse to fine details—much like a Matryoshka doll~\cite{cai2024matryoshka,kusupati2022matryoshka}.
The nested features capture structured information across multiple coarse-to-fine levels, enabling adaptable token lengths that align with different perceptual granularities.

\subsection{Denoising for Multimodal Brain Alignment} %
\label{subsec:alignment_strategy}

This section outlines training a brain encoder using features from~\cref{subsec:feature_space} as ground truth. 
Given brain responses $s \in \mathbb{R}^{1 \times L_s}$ (with $L_s$ indicating the dimension of input brain data) and associated visual stimuli $v \in \mathbb{R}^{W \times H \times C}$ (representing image width, height, and channels), we train the brain encoder $\mathcal{B}$ to minimize the distance between brain features $\mathbf{b}$ and target image features $\mathbf{v}$ from the vision encoder $\mathcal{V}$, aiming for a close approximation $\mathcal{B}(s) \approx \mathcal{V}(v)$. Toward the objective, the most straightforward alignment is through element-wise feature reconstruction~\cite{xia2024umbrae}:
\begin{equation}
\left.\mathcal{L}_\text{R}=\mathbb{E}_{\mathbf{b} \sim \mathbf{B}, \mathbf{v} \sim \mathbf{V}}[\| \mathcal{V}(v)-\mathcal{B}(s)) \|_2^2\right],  
\end{equation}
where the brain encoder $\mathcal{B}$ learns the alignment between source brain space $\mathbf{B}$ and target image space $\mathbf{V}$.

\begin{figure}
    \centering
    \includegraphics[width=0.78\linewidth]{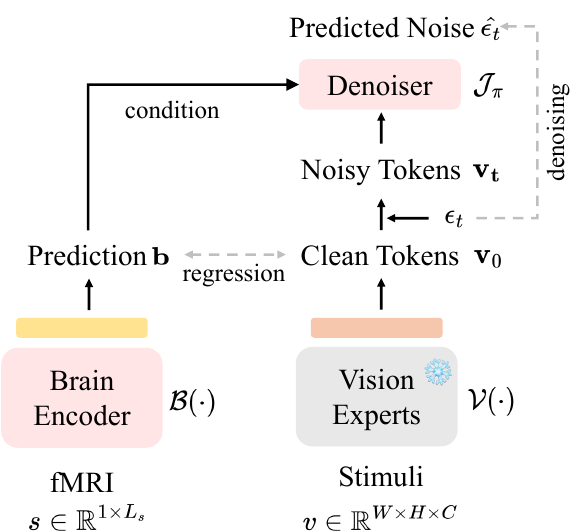}
    \vspace{-2mm}
    \caption{\textbf{Training Procedure with Denoiser.}}
    \label{fig:denoiser}
\end{figure}

\condparagraph{Regression or Denoising.} %
Regression using the Mean Squared Error (MSE, \ie, L2) loss directly learns a deterministic mapping from brain data points to those of images~\cite{xia2024umbrae}. However, one downside of this deterministic objective is that it can lead to overfitting, as the model may memorize specific instances rather than generalizing well to unseen data—especially considering the data scarcity in brain modalities. 
This issue becomes particularly problematic when dealing with noisy or incomplete data, as is often the case with brain signals, making the model sensitive to minor variations in the input.
Denoising~\cite{song2019generative,ho2020denoising,song2020denoising}, on the other hand, has been shown to be more effective than regression, as introducing noise into training data acts as an implicit form of data augmentation and regularization~\cite{wang2024reconstructive,li2025autoregressive}. The denoising process encourages the model to focus on the underlying data manifold rather than memorizing specific instance values~\cite{rombach2022high,song2019generative,ho2020denoising,song2020denoising}. 
Furthermore, visual signals suffer from heavy spatial redundancy~\cite{he2022masked}; recent work has shown that such redundancy also exists in visual features~\cite{cai2024matryoshka}.
Therefore, we introduce a masked denoising optimization objective as shown in~\cref{fig:denoiser}. The training procedure of the $\pi$-parameterized denoiser $\mathcal{J}_\pi$ follows a diffusion process:
\begin{equation}
  \mathcal{L}_{\text{D}} = \mathbb{E}_{\varepsilon, t} \left\|  \mathcal{J}_\pi \left( \mathbf{v}_t; \mathbf{b}, t \right) - \varepsilon \right\|^2.
\end{equation}
Here, $\varepsilon$ represents the ground truth noise sampled from the standard normal distribution $\mathcal{N}(\mathbf{0}, \mathbf{I})$. The noise-corruption is defined as $\mathbf{v}_t := \sqrt{\bar{\alpha}_t} \mathbf{v}+\sqrt{1-\bar{\alpha}_t} \varepsilon$, where $\bar{\alpha}_t$ is a noise schedule~\cite{ho2020denoising,nichol2021improved} at timestep $t$, and $\mathbf{v}=\mathcal{V}(v)$ maps input images $v$ to its representation. 
The noise estimator $\mathcal{J}_\pi\left(\mathbf{v}_t; \mathbf{b}, t\right)$ predicts the noise level in $\mathbf{v}_t$ at timestep $t$, conditioned on brain predictions $\mathbf{b}$. 
The positional embedding~\cite{li2025autoregressive} is added to the masked features to indicate the positions that need to be predicted, with the loss computed only on the unknown ones~\cite{he2022masked,li2025autoregressive}. The final loss as shown in~\cref{fig:denoiser} is the weighted sum of the regression loss and the denoising loss.

These features can be extracted from images on the fly during training or pre-calculated in advance to accelerate the process. During inference, the features predicted by the brain encoder, given brain signals as input, are fed into the remaining MLLM components (connector and LLM) for instruction-following tasks~\cite{liu2023visual,xia2024umbrae}, such as continuous dialogues, detailed descriptions, and complex reasoning.

\section{MG-BrainDUB}
\label{sec:benchmark}

To better evaluate the different choices of feature spaces in decoding fine-grained information, we propose a benchmark for detailed brain understanding. This involves curating high-quality evaluation datasets for two tasks: detailed captioning and salient question answering, each with corresponding metrics. We name this benchmark the Multi-Granularity Brain Descriptive Understanding Benchmark (MG-BrainDUB).

\subsection{Descriptive Caption Understanding}
\label{subsec:benchmark}

Current brain captioning methods~\cite{han2024onellm,xia2024umbrae,takagi2023improving,ferrante2023brain} face two primary challenges. First, they rely on COCO~\cite{lin2014microsoft} annotations, which provide only short captions with limited classes, lacking the granularity and vocabulary needed for detailed descriptions. Second, existing evaluation metrics are inadequate for assessing fine-grained captions. Traditional rule-based metrics, such as BLEU~\cite{papineni2002bleu}, CIDEr~\cite{vedantam2015cider}, and METEOR~\cite{banerjee2005meteor}, compute $n$-gram matching scores between candidate and reference captions, making them highly sensitive to stylistic variations and leading to inconsistent evaluations.
Model-based metrics, such as SPICE~\cite{anderson2016spice}, suffer from outdated text encoders as backbones, while CLIP-based alternatives~\cite{radford2021learning,hessel2021clipscore} enforce a strict 77-token input limit, further restricting their effectiveness in evaluating detailed captions.

We tackle both challenges through two key strategies. 
For detailed caption annotation, we construct ground truth descriptions using state-of-the-art MLLMs~\cite{liu2023visual}, followed by manual error correction, missing element addition, and hallucination removal through expert human intervention.
For evaluation metrics, in addition to leveraging CAPTURE~\cite{dong2024benchmarking}, a metric specifically designed for long captions, we introduce a structured evaluation framework that assesses captions based on precision, recall, and F1 scores of three core visual elements: objects, attributes, and relations~\cite{lu2024benchmarking,dong2024benchmarking}. To achieve this, we develop an automated pipeline with two key steps: visual element extraction and structured matching.

\condparagraph{(a) Element Extraction.}  Given detailed captions, we first split them into separate sentences using NLTK toolkit~\cite{loper2002nltk} and extract key visual elements—objects, attributes, and their relations—using a T5-based factual parser~\cite{raffel2020exploring,li2023factual}.

\condparagraph{(b) Structured Matching.} The matching process aligns extracted objects, attributes, and relations in three steps: exact matching (identifying precise word matches), synonym matching (matching words with similar meanings), and semantic matching (computing cosine similarity of word embeddings for any remaining unmatched elements). 

Finally, we compute the precision, recall, and F1 scores for visual elements based on the matching results. 
These metrics capture semantic discrepancies: precision penalizes hallucinations and recall captures omissions, reducing confounds from pretrained generative model biases and offering a more accurate measure of genuine brain-based decoding. %

\begin{table*}[t]
\caption{\textbf{Concept Localization.} We present brain grounding~\cite{xia2024umbrae} results derived directly from brain signals, reflecting spatial information encoded in brain activities. Image-based results are included as references, comprising visual grounding from Shikra~\cite{chen2023shikra} using associated visual stimuli and grounding performance based on reconstructed images from state-of-the-art visual decoding methods~\cite{ozcelik2023brain, scotti2023reconstructing,xia2024dream,xia2024umbrae}. 
We highlight the \tgold{best}, \tsilver{second-best}, and \tbronze{third-best}~performance averaged across all four subjects. 
}
\label{tab:brain_grounding}
\vspace{-2mm}
\centering
\setlength{\tabcolsep}{4pt}
\resizebox{0.8\textwidth}{!}
{
\begin{tabular}{@{}c|l|cc|cc|cc|cc|cc@{}}
\toprule
~ & \multirow{2}{*}{Method} & \multicolumn{2}{c|}{All} & \multicolumn{2}{c|}{Salient}  & \multicolumn{2}{c|}{Salient Creatures} & \multicolumn{2}{c|}{Salient Objects}  & \multicolumn{2}{c}{Inconspicuous}  \\ 
~ & ~ & acc@0.5  & IoU      & acc@0.5      & IoU      & acc@0.5       & IoU       & acc@0.5       & IoU       & acc@0.5       & IoU     \\
\midrule
\parbox[t]{3mm}{\multirow{5}{*}{\rotatebox[origin=c]{90}{Vision}}} & Shikra~\cite{chen2023shikra} & 51.96  & 47.22 & 62.92   & 56.44 & 66.71   & 59.34  & 58.79   & 53.27 & 38.29   & 35.71   \\
~ & BrainDiffuser~\cite{ozcelik2023brain} &       {17.49} &       {19.34} &       {27.18} &       {27.46} &       {38.71} &       {34.63} & 14.62  & 19.66 &       {5.39}  &       {9.20}   \\
~ & MindEye~\cite{scotti2023reconstructing} & 15.34 & 18.65 & 23.83 & 26.96 & 29.29  & 31.64 &       {17.88} &       {21.86} &       {4.74}  &       {8.28}  \\
~ & DREAM~\cite{xia2024dream} & {16.21} & 18.65 & {26.51} & 27.35 &       {34.43} &       {33.85} &       {17.88} & 20.28 & 3.35 & 7.78  \\
~ & UMBRAE~\cite{xia2024umbrae} &       {16.83} &       {18.69} &       {27.10} &       {27.55} & {34.14} & {33.65} &       {19.44} &       {20.92}  & 4.00 & 7.64 \\
\midrule
\parbox[t]{3mm}{\multirow{4}{*}{\rotatebox[origin=c]{90}{Brain}}} & UMBRAE-{S}~\cite{xia2024umbrae} & 14.31 & 18.13 & 22.28 & 25.82 & 26.32 & 29.58 & 17.88 & 21.72 & \tbronze{4.37} & 8.53 \\
~ & UMBRAE~\cite{xia2024umbrae} & \silver{18.03} & \silver{20.63} & \silver{28.22} & \silver{29.32} & \silver{36.93} & \silver{35.61} & \silver{18.74} & \silver{22.47} & \gold{5.32} & \silver{9.79} \\
~ & VINDEX-S & \bronze{14.92 }& \bronze{18.55} & \bronze{23.38} & \bronze{26.28} & \bronze{28.00} & \bronze{30.29} & \bronze{18.35} & \bronze{21.92} & \bronze{4.37} & \bronze{8.91} \\
~ & VINDEX & \gold{18.10} & \gold{20.89} & \gold{28.50} & \gold{29.78} & \gold{37.00}  & \gold{36.20} & \gold{19.25} & \gold{22.78} & \silver{5.11} & \gold{9.80} \\
\bottomrule
\end{tabular}
}
\end{table*}

\subsection{Question-Answering Annotation}

Previous studies have emphasized the multimodal evaluation for decoded \textit{salient} objects, considering semantic selectivities in the higher visual cortex of the human brain~\cite{kanwisher1997fusiform,desimone1984stimulus,puce1996differential}. Tiny objects in stimuli may not capture the subject's attention or relevant brain activities may not be effectively recorded during experiments~\cite{allen2022massive,xia2024umbrae}.
Thus, we further design evaluation focused on salient objects using annotated question-answering pairs. First, we extract salient objects using SAM~\cite{kirillov2023segment} and input the cropped regions into the MLLM to identify objects, attributes, and relationships within each image using pipelines introduced in~\cref{subsec:benchmark}.
We use SAM instead of the original COCO location annotations or Referring Expression Comprehension~\cite{chen2023shikra} with COCO classes~\cite{lin2014microsoft} to extract such information as pseudolabels, since COCO often overlooks salient categories outside its 80 common classes. For instance, an image may feature a large church, but only two small clocks on it are annotated with their spatial locations and linguistic descriptions.
For each identified object, we design questions such as ``which description best fits the \texttt{<obj>} in the image'' and ``what is the relationship between \texttt{<obj1>} and \texttt{<obj2>}' to evaluate the model’s ability to recognize objects and their interactions within the image. Additionally, we include in-depth reasoning questions~\cite{liu2023visual} that require step-by-step logical processing, based on GPT annotations~\cite{achiam2023gpt,shen2024neuro}, to assess deeper reasoning capabilities.

To mitigate the MLLM's tendency to default to affirmative responses such as ``Yes, I can see it'' rather than genuinely recognizing an object, we follow~\cite{lu2024benchmarking} and introduce an equivalent number of question-answer pairs for objects not present in the image as hallucination components. We then ask the model to select the correct description from multiple choices, where one option is a correct caption, and two are incorrect. The attribute annotations of fine-grained objects serve as the answers, while erroneous linguistic annotations are manually modified to ensure they contradict the actual visual content. We refer to this as salient QA (SQA) for complex reasoning and report the response accuracy.

\section{Experiments}
\label{sec:exp}

In our experiments, we demonstrate multimodal interaction with fMRI using MLLMs to tackle four tasks: concept localization (\cref{subsec:grounding}), concise captioning (\cref{subsec:short_caption}), descriptive captioning (\cref{subsec:detail_caption}), and complex reasoning (\cref{subsec:sqa}). While the generated multimodal explanations have been shown to enhance reconstruction performance~\cite{shen2024neuro,xia2024umbrae,takagi2023improving}, we prioritize brain perception over reconstruction.

\subsection{Implementation Details}

\condparagraph{Architecture.} 
For the mixture of vision encoders, we use pre-trained CLIP~\cite{radford2021learning} and DINOv2~\cite{oquab2023dinov2} as visual encoders. The target image features are extracted from both transformer encoders and then interleaved to spatially mix visual tokens before being subsequently processed by the connector and LLM~\cite{tong2024eyes}. The feature dimension $D$ is 4,096 for 7B models and 5,120 for 13B models.
For aggregated features, we utilize SigLIP~\cite{zhai2023sigmoid} to extract visual features from multiple layers~\cite{yao2025dense}. The connector first encapsulates multi-layer visual features and then maps the integrated visual features to the token space.
For nested features, we use CLIP~\cite{radford2021learning} to encode visual cues in a coarse-to-fine manner~\cite{cai2024matryoshka}, where an image is represented as patch-based visual tokens. We create $M$ sets of tokens, with the visual tokens at the coarser level derived directly from those at the finer level. Specifically, starting from the initial visual tokens, we sequentially apply $2 \times 2$ pooling with a stride of 2, resulting in $12 \times 12$, $6 \times 6$, and $3 \times 3$ visual tokens. Finally, we apply $3 \times 3$ pooling to obtain the most condensed single visual token.
Details on encoders and MLLMs can be found in the appendix.

\setlength{\tabcolsep}{4pt}
\setlength{\fboxrule}{0pt} 
\setlength{\fboxsep}{2pt}
\begin{table*}[t!]
\caption{\textbf{Brain Captioning}. We evaluate short captions on BrainHub~\cite{xia2024umbrae} and detailed captions on MG-BrainDUB. Models with `-{S}' refers to those with cross-subject support but are trained in a single-subject setting for comparison. MindEye2 results are based on a model adapted to one subject using a pre-trained model that was trained on the remaining seven subjects with the full dataset.
The mark$^{\dagger}$ denotes \textit{zero-shot} methods, indicating that no external captions were used during training.
`Shikra-img' refers to image captioning results from Shikra~\cite{chen2023shikra} using corresponding visual stimuli as input, acting as an approximate performance upper bound. 
Colors denote \tgold{best}, \tsilver{second-best}, and \tbronze{third-best} performance. 
\textbf{Bold} and \underline{underline} represent the best and second best results under identical protocols.
}
\vspace{-2mm}
\label{tab:brain_captioning}
\centering
\resizebox{0.8\textwidth}{!}
{
\begin{tabular}{@{}lcccccccccc@{}}
\toprule
Method & BLEU1 & BLEU2 & BLEU3 & BLEU4 & METEOR & ROUGE & CIDEr  & SPICE  & CLIP-S & RefCLIP-S\\
\midrule
& \multicolumn{10}{c}{Concise Captioning~(\cref{subsec:short_caption})} \\
\midrule
Shikra-img~\cite{chen2023shikra} & 82.38 & 69.90 & 58.63 & 49.66 & 35.60 & 65.49 & 161.43 & 27.62 & 80.60   & 85.92  \\
\midrule
SDRecon~\cite{takagi2023improving}  & 36.21 & 17.11 & 7.72 & 3.43 & 10.03 & 25.13  & 13.83 & 5.02 & 61.07 & 66.36   \\
OneLLM~\cite{han2024onellm} & 47.04 & 26.97 & 15.49 & 9.51 & 13.55 & 35.05  & 22.99 & 6.26 & 54.80    & 61.28   \\
BrainCap~\cite{ferrante2023brain} & {55.96} & {36.21} & {22.70} & {14.51} & 16.68 & {40.69} & {41.30} & {9.06} & {64.31} & {69.90} \\
NeuroVLA~\cite{shen2024neuro} & 57.19 & 37.17 & 23.78 & 15.85 & {18.60} & 36.67 & 49.51 & \bronze{12.39} & 65.49 & - \\
MindEye2~\cite{scotti2024mindeye2} & 54.82 & 38.60 & {26.49} & {18.16} & 17.54 & \bronze{43.77} & \bronze{55.70} & 10.97 & \bronze{67.54} & \bronze{73.73} \\
MEVOX~\cite{xia2025mevox}$^{\dagger}$ & \bronze{58.56} & \bronze{40.36} & \gold{28.09} & \gold{20.11} & \bronze{19.20} & \gold{44.47} & 54.37 & 11.05 & 64.23 & 70.36 \\
UMBRAE-S$^{\dagger}$ & {57.63} & {38.02} & {25.00}  & {16.76}   & {18.41}   &  {42.15}  & {51.93} & {11.83} & {66.44} & {72.12}   \\
UMBRAE~\cite{xia2024umbrae}$^{\dagger}$  & \silver{59.44} & \silver{40.48} & \silver{27.66} & \silver{19.03} & \silver{19.45} & {43.71}  & \gold{61.06} & \silver{12.79} & \silver{67.78} & \silver{73.54}  \\
\midrule
VINDEX-S$^{\dagger}$ & {57.99} & {39.00} & {26.04} & {17.83} & {18.56} & {42.27} & {53.96} & {11.92} & {66.94} & {72.64} \\ 
VINDEX$^{\dagger}$  & \gold{60.00} & \gold{40.72} & \bronze{27.57} & \bronze{18.91} & \gold{19.51} & \silver{43.78} & \silver{60.32} & \gold{12.84} & \gold{68.26} & \gold{73.88} \\
\midrule
& \multicolumn{10}{c}{Descriptive Captioning~(\cref{subsec:detail_caption})} \\
\midrule
NeuroVLA~\cite{shen2024neuro} & 38.91 & 24.02 & 15.24 & 12.41 & 18.44 & 27.83 & 42.58 & 18.41 & 56.16 & - \\
\midrule 
VINDEX &  & & & & & & & & & \\
\quad w/ SE & 22.71 & 12.89 & 7.21 & 4.11 & 13.95 & 25.43 & 14.05 & 8.43 & 62.90 & 67.72 \\
\quad w/ ME & 21.39 & 11.86 & 6.31 & 3.39 & 11.31 & 17.60 & 6.04 & 6.43 & 60.85 & 65.96 \\
\quad w/ AF & 0.09 & 0.01 & 0.00 & 0.00 & 0.04 & 1.89 & 0.03 & 0.01 & 53.17 & 59.86 \\
\quad w/ NF$_1$ & 20.10 & 11.32 & 6.14 & 3.40 & 14.67 & 21.15 & 1.47 & \underline{9.85} & 60.35 & 63.01 \\
\quad w/ NF$_9$ & \textbf{50.77} & \textbf{33.25} & \textbf{21.17} & \textbf{13.20} & \textbf{18.26} & \textbf{41.72} & \textbf{46.50} & \textbf{11.45} & \textbf{65.40} & \textbf{71.20} \\
\quad w/ NF$_{36}$ & \underline{38.87} & \underline{23.87} & \underline{14.39} & \underline{8.80} & \underline{14.72} & \underline{33.40} & \underline{31.07} & 8.94 & \underline{64.02} & \underline{69.80} \\
\quad w/ NF$_{144}$ & 3.74 & 2.22 & 1.27 & 0.74 & 3.46 &10.45 & 6.58 & 1.69 & 57.14 & 63.77 \\
\bottomrule
\end{tabular}
}
\end{table*}

\condparagraph{Training Details.} 
Our models are trained on a single A100 GPU. CLIP-224~\cite{radford2021learning} and DINO-224~\cite{oquab2023dinov2} feature alignment with a batch size of 128 takes approximately 8 hours, CLIP-336~\cite{radford2021learning} with a batch size of 64 takes about 23 hours, while SigLIP-384~\cite{zhai2023sigmoid} with a batch size of 4 requires around 40 hours to converge.
We use AdamW~\cite{loshchilov2017decoupled} as the optimizer with $\beta_1$=0.9, $\beta_2$=0.95, weight decay of 0.01, and apply the one-cycle strategy~\cite{smith2019super} with an initial learning rate of 3e-4 for the learning rate scheduler. 
Following visual decoding studies \cite{takagi2023improving,scotti2023reconstructing,xia2024dream}, we use the standard train and test splits for four subjects ({\small \texttt{sub01}, \texttt{sub02}, \texttt{sub05}, \texttt{sub07}}), each with  24,980 training samples and a shared 982 test samples. For evaluation, we report the average of the three repetitions of the same image in the test set, totaling 982 samples per subject. We implement the denoiser as an MLP with a default depth of 1, a latent dimension (width) of 1024, and a weight multiplier of 1.0 for the denoising loss (indicating the same contribution as the regression loss during training).

\subsection{Concept Localization}
\label{subsec:grounding}

This section evaluates the model's ability for concept localization through the brain grounding task~\cite{xia2024umbrae}. Similar to visual referring expression comprehension, brain grounding uses the prompt ``Find the coordinates of \texttt{<expr>}'' and evaluates performance using accuracy and IoU metrics. The accuracy metric, \texttt{acc@m}, measures the percentage of correctly labeled instances with an IoU greater than the threshold $m$. 
We assess grounding results derived directly from brain signals, with image-based results provided for reference. These include visual grounding using ground truth stimuli and reconstructed images from state-of-the-art (SOTA) visual decoding methods~\cite{ozcelik2023brain, scotti2023reconstructing, xia2024dream} via Shikra~\cite{chen2023shikra}.
We provide a comparison with SOTAs and our methods in~\cref{tab:brain_grounding}.

The results demonstrate that our method VINDEX outperforms all baselines across all metrics in the four salient-based categories in BrainHub~\cite{xia2024umbrae}. Notably, VINDEX surpasses both direct brain signal-based predictions from UMBRAE~\cite{xia2024umbrae} and image-based grounding results, which use reconstructed images from brain signals as input~\cite{xia2024umbrae, xia2024dream, scotti2023reconstructing, ozcelik2023brain}. This emphasizes that directly extracting location information from the captured brain activities is more accurate than relying on reconstructed images.
This observation suggests that reconstruction methods relying on image and text semantic encoding have inherent limitations in decoding spatial information. It also implies that incorporating grounding information in future models could help enhance the accuracy of object location prediction when reconstructing the visual stimuli from neural activations.

\setlength{\tabcolsep}{4pt}
\setlength{\fboxrule}{0pt} 
\setlength{\fboxsep}{2pt}
\begin{table*}[t!]
\caption{\textbf{Descriptive Captioning and Complex Reasoning}. We evaluate detailed descriptions and complex reasoning with salient question-answering on MG-BrainDub under different model settings. `Space' refers to the feature space elaborated in~\cref{subsec:feature_space}, while `size' and `encoder' indicate the size of ground truth images and the encoder used for feature extraction.
Colors denote \tgold{best}, \tsilver{second-best}, and \tbronze{third-best} performance. Models marked with $\ast$ correspond to those with the same settings as in~\cref{tab:brain_captioning}.
}
\vspace{-2mm}
\label{tab:brain_perception}
\centering
\resizebox{0.92\textwidth}{!}
{
\begin{tabular}{@{}llll|c|ccc|ccc|ccc|c|c@{}}
\toprule
\multicolumn{4}{c|}{Model Setting} & \multirow{2}{*}{CAPTURE} & \multicolumn{3}{c|}{Object}  & \multicolumn{3}{c|}{Attribute}  & \multicolumn{3}{c|}{Relation} & \multirow{2}{*}{Runtime (s)} & SQA \\
Space & Size & Encoder & MLLM &  & P & R & F1 & P & R & F1 & P & R & F1 & & Acc (\%) \\
\midrule
SE$^\ast$ & 224 & CLIP & LLaVA-1.5 7B & {0.4356} & 57.22 & 49.76 & 51.88 & 37.90 & 36.08 & 35.73  & 45.25 & 40.70 & 42.32  & 14.05 & 75.50 \\
SE & 224 & CLIP & LLaVA-1.5 13B  & \bronze{0.4733} & \bronze{61.81} & 52.91 & 55.60  & 40.68 & 41.21 & 39.38   & \silver{49.28} & 45.00 & \bronze{46.52} & 8.70  & 77.92 \\
SE & 224 & CLIP & LLaVA-1.6 7B  & 0.4684 & 51.02 & \silver{58.78} & 53.51 & 36.01 & \silver{54.01} & \gold{41.80}   & 41.00 & \bronze{45.63} & 42.75 & 10.91  & 76.80 \\
SE & 224 & CLIP & LLaVA-1.6 13B  & 0.4645 & 50.44 & \bronze{57.83} & 52.81 & 35.89 & \gold{54.02} & \silver{41.72}   & 40.42 & 45.44 & 42.38& 24.18  & \silver{78.69} \\
ME$^\ast$ & 224 & CLIP, DINO & LLaVA-1.5 13B  & 0.3041 & 49.17 & 34.30 & 38.07 & 26.10 & 23.09 & 23.37 & 33.08 & 26.86 & 28.86 & \bronze{5.06}  & 52.66 \\
AF$^\ast$ & 384 & SigLIP & LLaVA-1.5 7B & 0.0058 & 1.89 & 1.07 & 1.27 & 0.15 & 0.11 & 0.13 & 0.01 & 0.00 & 0.00 & 5.69  & 15.17 \\
NF$_1$$^\ast$ & 224 & CLIP & LLaVA-1.5 7B  & \silver{0.4877} & 59.19  & 57.19  & \silver{57.24} & \silver{42.74}  & 42.76  & \bronze{41.18}  & 47.32 & \silver{46.70}  & \silver{46.57} & 5.45 & \silver{82.59} \\
NF$_9$$^\ast$ & 224 & CLIP  & LLaVA-1.5 7B & \gold{0.5021} & \gold{62.43} & \gold{59.26} & \gold{59.74} & \bronze{41.97} & 43.20  & {40.95}  & \gold{50.56} & \gold{49.49} & \gold{49.54} & 5.31  &  \gold{83.83} \\
NF$_{36}$$^\ast$ & 224 & CLIP & LLaVA-1.5 7B & 0.4672 & \silver{62.23} & 54.89 & \bronze{56.37} & 40.24 & 38.56 & 37.58 & \bronze{47.90} & {44.41} & 45.47 & \silver{4.33} & 66.35 \\
NF$_{144}$$^\ast$ & 224 & CLIP & LLaVA-1.5 7B & 0.1614 & 52.84 & 22.47 & 28.73 & 8.67 & 5.55  & 6.32 & 12.88 & 7.67 & 9.22  & \gold{1.83}  & 31.35 \\
NF$_1$ & 224 & CLIP & LLaVA-1.6 7B & 0.4359 & 49.47  & 51.37  & 49.43  & 35.65  & \bronze{49.59} & 40.23  &36.75  & 38.96 & 37.40 & 8.62   & 65.58 \\
NF$_1$ & 336 & CLIP & LLaVA-1.5 7B & 0.4120 & 47.99 & 46.74 & 46.89 & \gold{43.01} & 39.50 & 39.53  & 31.15 & 31.96 & 31.16 &  \bronze{5.06} & 62.27 \\
\bottomrule
\end{tabular}
}
\end{table*}

\subsection{Concise Captioning}
\label{subsec:short_caption}

We evaluate short captions on BrainHub~\cite{xia2024umbrae}, demonstrating our method’s performance on multimodal language tasks.
\cref{tab:brain_captioning} presents an evaluation of our brain captioning compared to state-of-the-art baselines, including three subject-specific methods SDRecon~\cite{takagi2023improving}, BrainCap~\cite{ferrante2023brain}, OneLLM~\cite{han2024onellm}, and four cross-subject methods NeuroVLA~\cite{shen2024neuro}, MindEye2~\cite{scotti2024mindeye2}, MEVOX~\cite{xia2025mevox}, and UMBRAE~\cite{xia2024umbrae}. 
Our method, VINDEX, for concise captioning is built upon the same base MLLM Shikra~\cite{chen2023shikra} as in UMBRAE~\cite{xia2024umbrae} for a fair comparison. 
MindEye2 is first pre-trained on seven subjects and then adapted to the remaining subject using all available data.
NeuroVLA~\cite{shen2024neuro}, UMBRAE~\cite{xia2024umbrae}, and VINDEX are trained on four subjects instead of eight.
Caption results from Shikra~\cite{chen2023shikra} using associated visual stimuli for captioning act as an approximate upper bound.
VINDEX achieves top performance across most metrics for concise brain captioning, even surpassing methods trained with more subject data~\cite{scotti2024mindeye2} or external annotations~\cite{shen2024neuro}. The other brain-feature alignment methods UMBRAE~\cite{xia2024umbrae} and MEVOX~\cite{xia2025mevox} ranked the second and the third. This highlights the effectiveness of the cross-subject brain feature alignment strategy, showcasing superior zero-shot performance without the need for external training data and generalization across subjects without requiring separate models or subject-specific parameters.

NeuroVLA, MindEye2, UMBRAE, and VINDEX generate fluent, complete, and information-rich sentences, highlighting the advantages of using LLMs. Notably, VINDEX and UMBRAE achieve superior performance in a zero-shot setting, without relying on additional annotations, outperforming models trained with real captions from COCO~\cite{lin2014microsoft} or pseudo captions from MLLMs~\cite{liu2023visual}.
This demonstrates that directly aligning brain signals with image features offers a better way than using external visual, spatial, or linguistic annotations. It preserves more accurate semantic and spatial cues decoded from brain signals and avoids factual errors or stylistic inconsistencies from external annotations.

\subsection{Descriptive Captioning}
\label{subsec:detail_caption}

We present a quantitative comparison of descriptive captioning results for NeuroVLA~\cite{shen2024neuro} and our method with different feature space choices: single encoder (SE), mixture of encoders (ME), aggregated feature (AF), and nested features (NF$_n$) where the token size $n$ is selected from \{1, 9, 36, 144\} after downsampling.
The latter \cref{tab:brain_captioning} presents evaluations using regular caption metrics from~\cite{xia2024umbrae}.
Considering the limitations of traditional rule-based metrics (sensitive to writing style) and model-based metrics (unable to process long sequences), we evaluate descriptive captioning on MG-BrainDUB (\cref{subsec:benchmark}). 
Results in~\cref{tab:brain_perception} show that using NF$_9$ (with 9 visual tokens) as supervision for training brain encoders achieves the best performance on most metrics. 
Surprisingly, using dense feature aggregation (AF) instead yields the worst results, and models trained with vanilla CLIP features~\cite{radford2021learning} (SE setting at first four rows) remain inferior, even with larger and more sophisticated MLLMs (scaling from v1.5 to v1.6, 7B to 13B). 

Training with feature spaces from larger CLIP models (336 vs. 224) does not improve performance but significantly increases training time and computational costs.
The brain encoder aligned with CLIP-ViT-L-224 features (corresponding to 16$\times$16 tokens) generally transfers well to MLLMs designed with the CLIP-ViT-L-336 model (24$\times$24 tokens) as the visual encoder, often yielding better results.
AF performs the worst but with the largest token size (729 visual tokens for SigLIP-384~\cite{zhai2023sigmoid}). In contrast, aligning brain encoders with CLIP-224 requires only one-fifth of SigLIP-384's training time and one-third of CLIP-336's. 
Recall that NF downsamples the predicted CLIP features (the same features used in SE/ME) using pooling operations. Notably, 9-token alignment yields the best results, while 1-token alignment ranks second, suggesting that fewer tokens, though less expressive, align better with brain signals, which encode less information than real images. This also explains the inferior results for ME, where mixing brain-CLIP and brain-DINO features amplifies prediction imprecisions, leading to worse performance than using brain-CLIP features alone.

The performance drop with increasing feature magnitude suggests that, while richer image features benefit visual perception tasks~\cite{tong2024eyes,cai2024matryoshka,yao2025dense}, brain signals capture relatively less comprehensive visual elements. 
This may due to subjects' selective attention~\cite{puce1996differential,kanwisher1997fusiform} or limitations in capturing brain activities~\cite{allen2022massive,xia2024umbrae}. 
We expect additional features to enhance the perception of subtler elements, but they mainly introduce unnecessary hallucinations and increase training costs, as brain signals predominantly encode basic visual elements.

Results in~\cref{tab:brain_perception} also indicate that traditional caption metrics~\cite{papineni2002bleu,vedantam2015cider,banerjee2005meteor} are not reliable when scaling to longer, more detailed captions. The rule-based metrics show that ME has higher scores than NF$_1$, even though NF$_1$ predicts more accurate objects, attributes, and relations. CLIP-based metrics~\cite{hessel2021clipscore} also show deficiencies, as they present high similarities for all generated captions, even for AF, which in most cases lead to nonsensical outputs, such as garbled text and blank spaces. The inference time does not increase as expected with the model complexity and token length; instead, it exhibits irregularities. NF$_9$$^\ast$ presents the best results across most metrics while maintaining low computation cost in terms of inference time and memory consumption.

\subsection{Complex Reasoning}
\label{subsec:sqa}

We evaluate the model's ability to perform complex reasoning on salient objects using Salient Question Answering (SQA) (\cref{subsec:sqa}), with accuracy reported in~\cref{tab:brain_perception}. Unlike previous sections focused on visual content, SQA involves in-depth reasoning that demands step-by-step logical processing to provide the right answer.
This experiment highlights the model's ability to decode intricate characteristics of \textit{salient} elements, meaning the prominent objects most likely to capture the subject's attention during brain activity capture~\cite{allen2022massive}, which have been shown to be better preserved in brain data~\cite{xia2024umbrae}. 
From the table, NF$_9$$^\ast$ achieves the best SQA performance across image sizes, visual encoders, and MLLM settings, likely benefiting from a balance between rich visual features and brain data constraints, consistent with detailed description results.
In contrast, ME and AF sometimes produce meaningless outputs due to improper feature-brain signal alignment, caused by the added complexity of features during training. They also suffer from hallucinations, generating non-existent objects and misrepresenting attributes and relationships, resulting in accuracy rates at or below random guessing.

\section{Ablation Studies}
\label{sec:ablation}

\paragraph{Model-Agnostic Evaluation.} VINDEX is a versatile, model-agnostic brain decoding framework that can be easily adapted to various subjects, image sizes, visual encoders, MLLM settings, input prompts, and downstream tasks. Due to the architectural consistency of LLaVA-based variants~\cite{liu2023visual,yao2025dense,chen2023shikra,cai2024matryoshka,tong2024eyes}, our trained brain encoders can be directly integrated into different MLLMs without requiring additional modifications. Specifically, for model setups presented in~\cref{tab:brain_perception}, we only need to train four brain encoders using CLIP-224, CLIP-336, DINO-224, and SigLIP-384. The same brain encoder trained with CLIP-224 can be used across SE, ME, and NF (except for the last one with a 336 image size). This brain-CLIP features are then passed to the remaining MLLMs components, including connectors and LLMs, with task prompts supported by the base MLLM for multimodal interactions, such as interactive dialogue, brain captioning, complex reasoning, and concept localization. The brain encoder aligned with CLIP-224 features adapts well to LLaVA-1.6-7B/13B models, which are designed to use CLIP-336 as the visual encoder, leading to improved performance. This versatility enables our method to efficiently align brain signals with multiple MLLMs at minimal cost. %

\condparagraph{Denoising Optimization Objective.} We conduct ablation studies on the denoiser in~\cref{tab:ablation_denoiser} to evaluate the effectiveness of the denoising objective over vanilla regression across different settings, focusing on architectural depth $d$, width $w$, and the weight multiplier $\beta$. Here, `S$n$' represents different configurations of specific depth, width, and weight multiplier; S0 refers to the base setup with only the regression loss, without the denoising loss. The results are reported as CAPTURE scores~\cite{dong2024benchmarking} (C) for detailed descriptions across each setup. Total training hours for 300 epochs across all settings is approximately eight hours, with no notable variations. We found that a simple one-depth MLP with 1,024 model channels yields the best results.
Results show that training with the denoising loss setup incurs no extra cost but leads to improved performance and more stable training. 

\setlength{\tabcolsep}{4pt}
\begin{table}[t!]
\centering
\caption{\textbf{Ablation Study on Denoiser}. CAPTURE~\cite{dong2024benchmarking}
is reported across architectural depth $d$, width $w$, and weight multiplier $\beta$.}
\label{tab:ablation_denoiser}
\resizebox{\linewidth}{!}
{
\begin{tabular}{ccccccccc}
\toprule
{Setup} & S0 & S1 & S2 & S3 & S4 & S5 & S6 & S7 \\ 
\midrule
{$\beta$} & - & 1.0 & 1.0 & 1.0 & 1.0 & 0.5  & 1.5 & 2.0 \\ 
{$d$} & - & 1 & 2 & 3 & 1 & 3 & 3 & 3  \\ 
{$w$} & - & 512  & 1,024 & 1,024 & 1,024  & 1,024 & 1,024 & 1024 \\ 
\midrule
C &  0.4179 & \underline{0.4308} & 0.4257 & 0.4183 & \textbf{0.4356} & 0.4193 & 0.4197 & 0.4096 \\ 
\bottomrule
\end{tabular}
}
\end{table}

\section{Conclusion}
\label{sec:conclusion}

In this paper, we propose a zero-shot multimodal brain decoding method that explores representative feature spaces from the visual components within MLLMs. We elucidate different vision feature space choices and introduce a denoiser-based strategy to enhance brain alignment with pre-trained models, improving the accuracy and robustness of brain-encoded feature representations.
Our method facilitates efficient and flexible decoding across multiple levels of granularity, eliminating the need for additional textual or spatial annotations during training. 
Furthermore, we present MG-BrainDUB, a benchmark aimed at providing a more reliable and comprehensive evaluation for multigranular multimodal brain decoding. It comprises two core tasks—detailed description generation and salient question answering—along with a novel metric that prioritizes key visual elements including objects, attributes, and relationships, offering a robust framework for assessing fine-grained brain perception.

\vspace{.35em}\noindent{}{\scriptsize%
\textbf{Acknowledgements.} This work was supported by a UKRI Future Leaders Fellowship [grant number G104084].\par%
}

\appendix
\onecolumn
\section*{Appendix}

This document includes further analyses on the background knowledge, experiments, and results.
We first provide more details Natural Scenes Dataset, vision encoders, and diffusion models in~\cref{sec:supmat_background}. 
\cref{sec:supmat_implementation} provides more implementation details in network architecture and prompt templates. 
We then detail the benchmark in \cref{subsec:supmat_benchmark} and provide additional results, analysis, and visualizations in \cref{sec:supmat_more_results}.
We discuss limitations and future directions, including possible choices of visual encoders and hybrid multimodal large language models, in \cref{sec:supmat_discussion}.

\section{Background}
\label{sec:supmat_background}

\subsection{NSD Dataset}
\label{subsec:supmat_nsd_data} 

We use the Natural Scenes Dataset (NSD)~\cite{allen2022massive} for the experiment. NSD is currently the largest released fMRI dataset, featuring detailed brain activity recordings from 8 subjects who passively viewed images sourced from the Common Objects in Context (COCO) dataset~\cite{lin2014microsoft} for up to 40 hours in an MRI machine. Each image was displayed for three seconds and repeated three times over 30-40 scanning sessions, resulting in 22,000-30,000 fMRI response trials per participant.

We follow the data preprocessing procedure similar to prior brain visual decoding studies~\cite{takagi2023improving,xia2024dream,scotti2023reconstructing,xia2024umbrae,scotti2024mindeye2} based on NSD~\cite{allen2022massive}. Specifically, we use preprocessed fMRI voxels in a 1.8-mm native volume space that corresponds to the ``nsdgeneral'' brain region. This region is described as the subset of voxels in the posterior cortex that are most responsive to the presented visual stimuli~\cite{allen2022massive}.
We train our model on the four subjects (with IDs 1, 2, 5, and 7) who completed all scanning sessions. The training set for each subject consists of 8,859 images and 24,980 fMRI trials, with each image shown up to three times. The remaining 982 images and 2,770 fMRI trials, which are common across all four participants, are used for testing. For fMRI data spanning multiple trials, we calculate the average response as in prior research~\cite{scotti2023reconstructing}.
\cref{tab:supmat_nsd_dataset} details characteristics of NSD and region of interests (ROIs) included in the fMRI data.

\subsection{Vision Encoders}
\label{subsec:supmat_vision_encoders}

\paragraph{CLIP.} CLIP~\cite{radford2021learning}, a contrastive language-image pre-training method, has gained significant attention for leveraging softmax contrastive learning on large-scale image-text datasets. As a contrastively pre-trained model, CLIP is widely used in various downstream applications, generating diverse representations for tasks such as object detection and semantic segmentation, and it demonstrates strong performance in zero-shot transfer tasks, including classification and retrieval. It is one of the most popular visual encoders in vision-language models and multimodal large language models, serving as the visual component~\cite{liu2023visual,yao2025dense,cai2024matryoshka}.

\paragraph{DINO.} DINO~\cite{oquab2023dinov2} is a self-supervised learning framework known for its ability to learn high-quality visual representations without relying on labeled data. Built on the Vision Transformer (ViT)~\cite{dosovitskiy2020image} and utilizing knowledge distillation~\cite{hinton2014distilling}, DINO effectively captures semantic structures in images. It has been widely applied to various computer vision tasks, including object detection, semantic segmentation, and visual grounding. DINO's strong feature representations make it a valuable visual encoder in multimodal contexts, enhancing spatial understanding~\cite{tong2024eyes,chung2025unifying,shen2025mome,zong2024mova}.

\paragraph{SigLIP.} SigLIP~\cite{zhai2023sigmoid} is a contrastive language-image pre-training framework designed to learn high-quality visual representations using sigmoid loss. Building on the foundation of CLIP, SigLIP incorporates a memory-efficient architecture and optimization strategies that enhance performance. The pre-trained SigLIP also serves as a feature extractor in the visual component of multimodal large language models~\cite{yao2025dense}.

\subsection{Diffusion Models}
\label{subsec:supmat_ldm}

The diffusion models~\cite{ho2020denoising,song2020denoising,song2019generative,nichol2021improved} typically comprise a forward process and a corresponding reverse process. The forward process adds noise, while the reverse denoising process learns to remove it. The model can operate in either pixel space~\cite{ho2020denoising} or latent space~\cite{rombach2022high}.
For latent diffusion model, given clean latent tokens $\bm{z}_0$ drawn from $p(\bm{z})$, the forward diffusion process is a Markov chain that performs progressive noise addtion to the original sample:
\begin{equation}
    q(\bm{z}_t | \bm{z}_{t-1}) = \mathcal{N}(\sqrt{1 - \beta_t} \bm{z}_{t-1}, \beta_t\mathbf{I}), 
\end{equation}
where $\mathcal{N}(\bm{\mu}, \bm{\Sigma})$ denotes the prior Gaussian distribution, $\beta_t \in (0, 1)$ indicates a pre-defined time-dependent variance schedule at discrete timestep $t$.
For sampling $\bm{z}_t$ from $\bm{z}_0$ at an arbitrary timestep $t$~\cite{ho2020denoising}, this can be reformulated as 
\begin{equation}
\begin{aligned}
\label{eq:add_noise}
    &q(\bm{z}_t | \bm{z}_0) = \mathcal{N}(\sqrt{\bar{\alpha}_t} \bm{z}_0, (1 - \bar{\alpha}_t) \mathbf{I}), \\
    &\bm{z}_t = \sqrt{\bar{\alpha}_t} \bm{z}_0 + \sqrt{1 - \bar{\alpha}_t} \bm{\epsilon}, \quad \bm{\epsilon} \sim \mathcal{N}(\bm{0}, \mathbf{I}),
\end{aligned}
\end{equation}
where $\alpha_t = 1 - \beta_t$ and $\bar{\alpha}_t = \prod_{i=1}^t \alpha_t$.
The reverse process in the latent diffusion model learns to denoise the added noise for latent tokens.
The reverse process iteratively generates clean tokens $\bm{z}_0$ from pure noise $\bm{z}_T$ conditioned on $\mathcal{C}$, as described by 
\begin{equation}
    \bm{z}_{t-1} = \frac{1}{\sqrt{\alpha_t}} \left( \bm{z}_t - \frac{1 - \alpha_t}{\sqrt{1 - \bar{\alpha}_t}} \bm{\epsilon}_{\pi}(\bm{z}_t; \mathcal{C}, t) \right) + \sigma_t \bm{\epsilon},
\end{equation}
where a $\pi$-parameterized denoiser $\bm{\epsilon}_{\pi}$ is trained to predict the added noise during the forward process and $\sigma_t$ indicates the posterior noise variance.
The objective for training the denoiser $\bm{\epsilon}_{\pi}$ is 
\begin{equation}
    \label{eq:dm}
    \mathcal{L}(\pi, \bm{z}_0) = \mathbb{E}_{t, \bm{\epsilon}} \left[
    || \bm{\epsilon}_{\pi} (\sqrt{\bar{\alpha}_t}\bm{z}_0 + \sqrt{1 - \bar{\alpha}_t} \bm{\epsilon}; \mathcal{C}, t) - \bm{\epsilon} ||^2
    \right].
\end{equation}

\begin{table}
\centering
\caption{\textbf{Details on NSD.} This table presents the training and test image-fMRI pairs for the four subjects, along with ROIs.}
\label{tab:supmat_nsd_dataset}
 \resizebox{0.55\linewidth}{!}
 {
    \begin{tabular}{ccccc}
        \toprule
        Training & Test & ROIs & Subject ID & Dimension \\ 
        \midrule
        \multirow{4}{*}{8,859} & \multirow{4}{*}{982} & \multirow{4}{*}{\begin{tabular}[c]{@{}c@{}}V1, V2, V3, hV4,\\ VO, PHC, MT,\\ MST, LO, IPS\end{tabular}} & \texttt{sub01} & 15,724 \\ 
                  &              &      & \texttt{sub02}      & 14,278   \\ 
                  &              &      & \texttt{sub05}      & 13,039   \\ 
                  &              &      & \texttt{sub07}      & 12,682   \\ 
        \bottomrule
    \end{tabular}
 }
\end{table}

\section{Implementation Details}
\label{sec:supmat_implementation}

\subsection{Architecture}

\begin{figure}[th]
    \centering
    \includegraphics[width=0.75\linewidth]{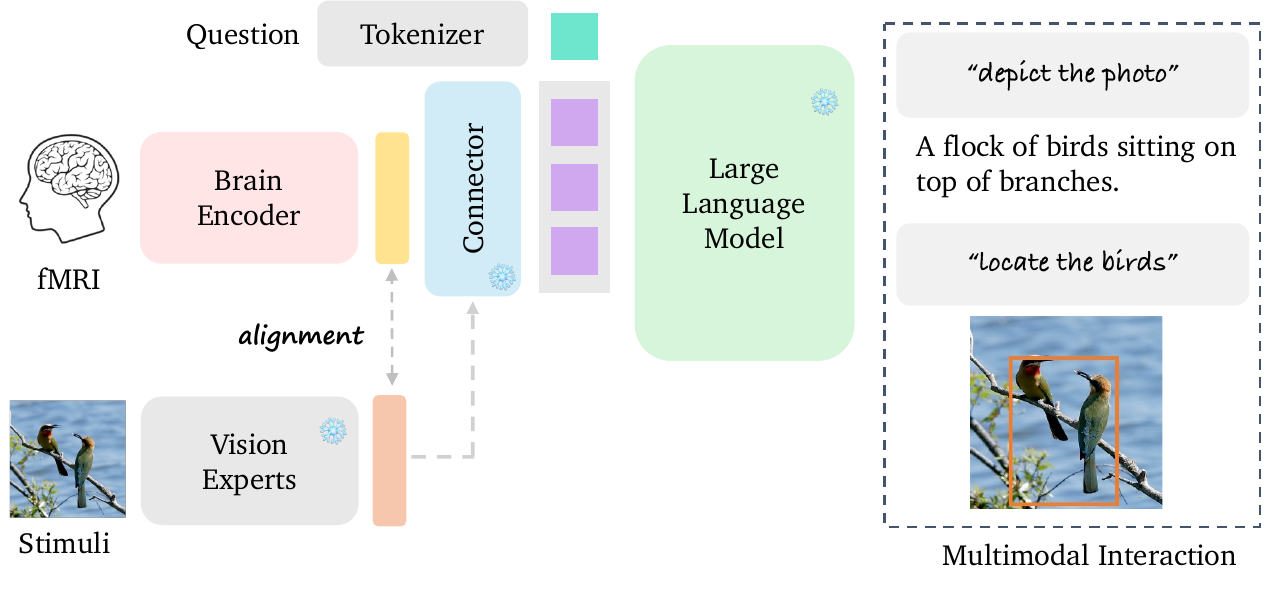}
    \caption{\textbf{Method Overview.} Once a pre-trained brain encoder is available, brain signals can be input to obtain predicted brain features. These brain features are then fed into the connector and LLM for multimodal brain interaction. Our method follows a similar overview to UMBRAE~\cite{xia2024umbrae} but differentiates itself through the use of different vision encoders, alignment strategies, and support for additional tasks.
    }
    \label{fig:supmat_overview}
\end{figure}

Our method includes a brain encoder and a multimodal large language model (MLLM). The features from visual encoders, as visual components of the MLLM, are used to train the brain encoder, allowing it to learn to predict brain features from the input brain signals. During inference, given a brain signal as input, the brain encoder predicts brain features, which replace the image features and are fed into the MLLM for instruction-following tasks. The overview is shown in~\cref{fig:supmat_overview}.

\paragraph{Brain Encoder.} For the brain encoder, we follow the UMBRAE architecture~\cite{xia2024umbrae}, which has shown to be an effective model for brain-image alignment. This encoder contains subject-specific tokenizers to handle individual subject information and a shared pre-trained trunk to capture the common information across subjects. We train the brain encoder to align the image features from the chosen vision encoder, allowing us to integrate brain features into the MLLM for brain perception tasks. Details on the trained brain encoder with different vision encoder settings and the supported MLLMs are provided in \cref{tab:pre-trained_brain_encoder}.

\begin{table}[th]
    \centering
    \renewcommand{\arraystretch}{1.2} 
    \caption{\textbf{Detailed on Our pre-trained Brain Encoders.} Each pre-trained brain encoder is trained to align with vision features from a target vision encoder, using a specific image size. The predicted brain features have a shape of (batch size, tokens, dimension). 
    The same single pre-trained brain encoder can support multiple off-the-shelf MLLMs. Notably, the trained brain encoders \textsc{b-clip224} and \textsc{b-clip336} are directly compatible with all eleven LLaVA models~\cite{liu2023visual} using different setting available in the \href{https://github.com/haotian-liu/LLaVA/blob/main/docs/MODEL_ZOO.md}{MODEL ZOO}, without any etra modifications.}
    \label{tab:pre-trained_brain_encoder}
    \resizebox{\linewidth}{!}
    {
    \begin{tabular}{llcccl}
    \toprule
    Brain Encoder & Vision Encoder & Size & \#Token &\#Dim & Supported MLLMs\\ 
    \midrule
    \textsc{b-clip224} & CLIP-224~\cite{radford2021learning} & 224 & 256 & 1024 & LLaVA-1.5/1.6-7B/13B, LLaVa-MoF-7B/13B~\cite{tong2024eyes}, M3-1.5/1.6-7B/13B~\cite{cai2024matryoshka}, Shikra~\cite{chen2023shikra} \\
    \textsc{b-clip336} & CLIP-336~\cite{radford2021learning} & 336 & 576 & 1024 & LLaVA-1.5/1.6-7B/13B, LLaVa-MoF-7B/13B~\cite{tong2024eyes}, M3-1.5/1.6-7B/13B~\cite{cai2024matryoshka}, DC~\cite{yao2025dense} \\
    \textsc{b-dino224} & DINOv2~\cite{oquab2023dinov2} & 224 & 256 & 1024 & LLaVa-MoF~\cite{tong2024eyes}\\
    \textsc{b-siglip384}& SigLIP-384~\cite{zhai2023sigmoid} & 384 & 729 & 1152 & DC~\cite{yao2025dense} \\
    \bottomrule
    \end{tabular}
    }
\end{table}

\paragraph{Vision Encoder.} The vision encoders are used to extract features for training the brain encoder. The vision encoders, as the visual component in MLLMs, are versatile, considering the architecture-conscious LLaVA-based MLLMs~\cite{liu2023visual,cai2024matryoshka,yao2025dense,tong2024eyes}.
We use four vision encoders, which act as representative feature spaces to demonstrate our brain alignment idea: CLIP-224 (\textsc{openai/clip-vit-large-patch14})~\cite{radford2021learning}, CLIP-336 (\textsc{openai/clip-vit-large-patch14-336})~\cite{radford2021learning}, DINOv2 (\textsc{facebookresearch/dinov2})~\cite{oquab2023dinov2}, and SigLIP-384 (\textsc{google/siglip-so400m-patch14-384})~\cite{zhai2023sigmoid}. The same brain encoder trained with CLIP-224/336 can be used across SE, ME, and NF. DINOv2 is used in ME, while SigLIP is used in AF.

\paragraph{MLLMs.} Our method supports various MLLMs with different configurations. An MLLM consists of a vision encoder, a connector, and a base LLM. Once a pre-trained brain encoder is available, brain signals can be input to obtain predicted brain features. As shown in~\cref{fig:supmat_overview}, these brain features are then fed into the connector and LLM for interaction. \cref{tab:pre-trained_brain_encoder} presents MLLMs supported by each trained brain encoder. Taking \textsc{b-clip224} as an example, for a batch of \texttt{bs} input brain signal from NSD~\cite{allen2022massive}, it produces brain-CLIP features of size (\texttt{bs}, 256, 1024). The connector then projects these features to (\texttt{bs}, 256, 4096) or (\texttt{bs}, 256, 5120), depending on whether the LLM is 7B or 13B, respectively. The same tokens, after projection, can be fed into the corresponding LLM or, in the NF setting, downsampled to 144, 36, 9, or 1 token for M3~\cite{cai2024matryoshka} processing.
In the AF setting, the predicted brain features have a size of (\texttt{bs}, 729, 1152). After aggregation from dense features across different layers, the size becomes (\texttt{bs}, 729, 3456) before being fed into the DC~\cite{yao2025dense} connector and LLM.

\paragraph{Denoiser.} 
For the denoising network, we use a small MLP following~\cite{ho2020denoising,song2019generative,li2025autoregressive} with several residual blocks~\cite{he2016deep}. Each sequentially applies a LayerNorm (LN)~\cite{ba2016layer}, two linear layers with SiLU activation in between, and merges with a residual connection. The denoising MLP is conditioned on brain predictions $\mathbf{b}$. The prediction $\mathbf{b}$ is added to the time embedding of the noise schedule at timestep $t$, serving as the condition for the MLP in LN layers.
The diffusion process follows~\cite{nichol2021improved}. The noise schedule has a cosine shape, with 1,000 steps at training time. The denoising network predicts the noise vector $\epsilon$~\cite{ho2020denoising}.

Our denoiser differs from the implementation of diffusion prior~\cite{scotti2023reconstructing} in the following aspects: (1) training target: The diffusion prior is applied to CLIP embeddings, whereas ours is applied to features from different visual encoders; (2) training effect: diffusion prior predicts CLIP embeddings through additional operations, while ours is only used for training the denoiser and not during prediction. Given that both features and images have inherent redundancy, we apply a mask, which acts as an implicit form of data augmentation and regularization; (3) structure: our denoiser is more lightweight in comparison.

\subsection{Prompt Template}

The choice of feature spaces in vision encoders for our model, including SE, ME, AF, and NF, follows similar LLaVA-based architectures and training procedures~\cite{liu2023visual}. Therefore, prompt templates for brief and detail image  description~\cite{liu2023visual,yao2025dense,cai2024matryoshka} used in LLaVA~\cite{liu2023visual}, as illustrated in \cref{tab:supmat_concise_describe_instructions} and \cref{tab:supmat_detailed_describe_instructions}, can be directly used as instructions for our concise and descriptive brain captioning. It should be noted that all these prompt templates are only used for inference in our experiments and are not utilized for generating training data.
For our method built upon Shika~\cite{chen2023shikra}, we use the prompts ``Describe the image \texttt{<image>} as simply as possible'' for concise captioning and ``Locate \texttt{<expr>} in \texttt{<image>} and provide its coordinates'' for concept localization, where \texttt{<expr>} represents the target expression and \texttt{<image>} serves as a placeholder for image features. The full prompt template, including the system message, user prompt, and assistant answers, follows:

\begin{tcolorbox}[left=0.2em, right=0.2em, top=0.2em, bottom=0.2em,colback=white,colframe=black!30!white]
    \centering
    \small
    \texttt{SYSTEM MESSAGE.}
    USER: \texttt{<image>} \textcolor{VioletRed}{\texttt{<instruction>}} 
    ASSISTANT: \textcolor{DarkOrchid}{\texttt{<answer>}}
\end{tcolorbox}
\noindent{}The tags \textcolor{VioletRed}{\texttt{<instruction>}} and \textcolor{DarkOrchid}{\texttt{<answer>}} serve as placeholders for human instructions and assistant answers.
We use variable templates for different tasks. Prompts for interactive dialogue and complex reasoning can be found in~\cite{liu2023visual}.

\begin{table*}[th!]
\centering
\caption{\textbf{Prompt for Concise Brain Captioning.} The list of instructions present the same meaning with natural language variance.} %
\label{tab:supmat_concise_describe_instructions}
\begin{minipage}{0.9\columnwidth}\vspace{0mm}    \centering
\begin{tcolorbox} 
\centering
\small
\hspace{-5mm}
\begin{itemize}[leftmargin=7.5mm]
\setlength{\itemsep}{2pt}
\item ``Describe the image concisely.''
\item ``Provide a brief description of the given image.''
\item ``Offer a succinct explanation of the picture presented.''
\item ``Summarize the visual content of the image.''
\item ``Give a short and clear explanation of the subsequent image.''
\item ``Share a concise interpretation of the image provided.''
\item ``Present a compact description of the photo's key features.''
\item ``Relay a brief, clear account of the picture shown.''
\item ``Render a clear and concise summary of the photo.''
\item ``Write a terse but informative summary of the picture.''
\item ``Create a compact narrative representing the image presented.''
\end{itemize}
\end{tcolorbox}
\end{minipage}
\end{table*}

\begin{table*}[h!]\centering
\caption{\textbf{Prompt for Descriptive Brain Captioning.} The list of instructions present the same meaning with natural language variance.}
\label{tab:supmat_detailed_describe_instructions}
\begin{minipage}{0.9\columnwidth}\vspace{0mm}    \centering
\begin{tcolorbox} 
\centering
\small
\hspace{-5mm}
\begin{itemize}[leftmargin=7.5mm]
\setlength{\itemsep}{2pt}
    \item ``Describe the following image in detail''
    \item ``Provide a detailed description of the given image''
    \item ``Give an elaborate explanation of the image you seev
    \item ``Share a comprehensive rundown of the presented image''
    \item ``Offer a thorough analysis of the image''
    \item ``Explain the various aspects of the image before you''
    \item ``Clarify the contents of the displayed image with great detail''
    \item ``Characterize the image using a well-detailed description''
    \item ``Break down the elements of the image in a detailed manner''
    \item ``Walk through the important details of the image''
    \item ``Portray the image with a rich, descriptive narrative''
    \item ``Narrate the contents of the image with precision''
    \item ``Analyze the image in a comprehensive and detailed manner''
    \item ``Illustrate the image through a descriptive explanation''
    \item ``Examine the image closely and share its details''
    \item ``Write an exhaustive depiction of the given image''
\end{itemize}
\end{tcolorbox}
\end{minipage}
\end{table*}

\section{Details on MG-BrainDUB}
\label{subsec:supmat_benchmark}

We describe in the main paper a benchmark for evaluating detailed captions, including annotations and metrics, as well as salient question answering (QA) for complex reasoning. Here, we provide examples to elaborate the evaluation process (\cref{subsec:supmat_metric_explanation}) and showcase constructed exemplars for detailed captioning (\cref{subsec:supmat_example_caption}) and salient QA (\cref{subsec:supmat_example_qa}).

\subsection{Metric Calculation Explanation}
\label{subsec:supmat_metric_explanation}

We demonstrate in experiments the drawbacks of current traditional rule-based and model-based metrics, necessitating the introduction of new metrics that consider visual elements for long, detailed caption evaluation. Our metric extracts and matches each caption based on precision, recall, and F1 scores for three core visual elements: objects, attributes, and relations~\cite{lu2024benchmarking,dong2024benchmarking}.%
To understand the evaluation process, we break down the metric calculation into steps. Give two example captions as the candidate and reference descriptions, we parse the description into a list containing tuples of objects, attributes, and relations, following the steps below:

\vspace{2mm}
\begin{itemize}[leftmargin=17.5mm]
\setlength{\itemsep}{2pt}
\small
{
\item[Candidate] ``A red car and a white truck are driving down a city street lined with green trees. Tall buildings in the background.''
\item[Reference] ``A peaceful beach with soft white sand stretching along the coastline, where turquoise ocean waves gently roll onto the shore. Several people are sunbathing near the water while others are playing volleyball in the distance.''
}
\end{itemize}
\vspace{2mm}

\condparagraph{Object Extraction.} This step identifies and lists all entities or objects mentioned in the description. For the given candidate caption, we get a list of object labels: 
[`building', `city street', `truck', `tree', `car'].

\condparagraph{Attribute Mapping.} This step identifies attributes associated with each object, which describe their properties or characteristics. The attribute mapping for the caption is a dictionary mapping object labels to their attributes as follows:
\{`car': \{`red'\}, `tree': \{`green'\}, `truck': \{`white'\}, `building': \{`tall'\}\}.
Each object is paired with its respective attributes, providing essential information for evaluating the model's ability to recognize both the objects and their attributes.

\condparagraph{Relation Extraction.} This discerns the relationships between different objects in the scene, which describe their spatial or functional connections. In the example caption, the relationships are:
\{(`truck', `drive down', `city street'), (`car', `drive down', `city street'), (`building', `in', `background')\}.
This information is essential for evaluating the model's ability to reason and represent spatial relationships accurately in the caption.

The structured results for the candidate and reference captions are as follows. This hierarchical data structure aids in evaluating the model's ability to recognize objects, their attributes, and the relationships between them in the scene.

\vspace{2mm}
\begin{itemize}[leftmargin=17.5mm]
\setlength{\itemsep}{2pt}
\small
{
\item[Candidate] objects: `building', `city street', `truck', `tree', `car'

attributes:  `car': {`red'}, `tree': {`green'}, `truck': {`white'}, `building': {`tall'},

{relations}:  
(`truck', `drive down', `city street'), 
(`car', `drive down', `city street'), 
(`building', `in', `background')

\item[Reference] {objects}:  `sky', `edifice', `car', `street', `truck', `city street', `tree'

{attributes}: `car': {`red'},
  `city street': {`busy'},
  `truck': {`white'},
  `tree': {`green'},
  `edifice': {`modern'},
  `sky': {`blue', `clear'}

{relations}: (`tree', `surround', `street'), 
 (`edifice', `stand under', `sky'), 
 (`car', `run in front of', `city street')
}
\end{itemize}
\vspace{2mm}

Once the elements are extracted from both the ground truth and candidate captions, we can compute the scores for objects, attributes, and relationships using the following metrics: 
(a) \textbf{Precision}, which measures the accuracy of the model on all mentioned samples in the candidate; 
(b) \textbf{Recall}, which measures the accuracy on all actual samples in the reference; 
(c) \textbf{F1 score}, which combines precision and recall by representing their harmonic mean.
Similar to the long detailed caption evaluation~\cite{lu2024benchmarking,dong2024benchmarking}, the scores for objects, attributes, and relationships are calculated as follows:
\begin{equation}
\text{Precision} = \frac{N(\text{Matched})}{N(\text{Candidate})}, \quad
\text{Recall} = \frac{N(\text{Matched})}{N(\text{Reference})}, \quad
\text{F1} = \frac{2 \cdot \text{Precision} \cdot \text{Recall}}{\text{Precision} + \text{Recall}},
\end{equation}
where \( N(\text{Matched}) \) is the number of correctly matched items, \( N(\text{Candidate}) \) the total items in the candidate, and \( N(\text{Reference}) \) the total in the reference.

Given that objects, attributes, and relationships extracted from ground truth and candidate captions are often not the same, we process the data through three matching steps: (a) {exact matching}: this step checks for precise word matches between the ground truth and candidate captions,
(b) {synonym matching}: this step matches words based on their similar meanings,
(c) {semantic matching}: for any remaining unmatched elements, the cosine similarity of their word embeddings is computed to determine their relevance. Using the same example, `city street', `truck', `tree', and `car' pass exact matching, while `building' passes synonym matching as it shares a similar meaning with `edifice'. This gives 5 correct matches and 2 missing matches. Therefore, the precision is 100\%, which is calculated as 5/5, recall is 71.43\% (5/7) and F1 is 83.33\%.

\subsection{Detailed Caption Examples} 
\label{subsec:supmat_example_caption}

\begin{figure}[t!]
\centering
\renewcommand{\arraystretch}{0.8}
\setkeys{Gin}{width=0.15\linewidth}
\setlength{\tabcolsep}{1.2pt}
    \begin{tabular}{ccccc}
         \subfloat{\includegraphics[width=0.18\textwidth]{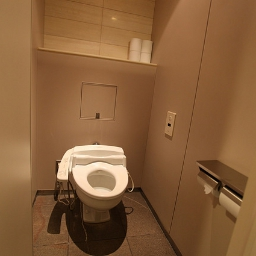}} &
         \subfloat{\includegraphics[width=0.18\textwidth]{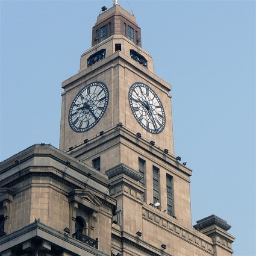}} &
         \subfloat{\includegraphics[width=0.18\textwidth]{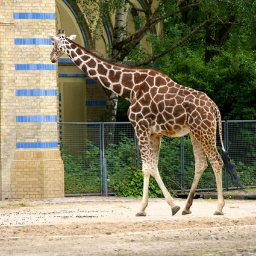}} &
         \subfloat{\includegraphics[width=0.18\textwidth]{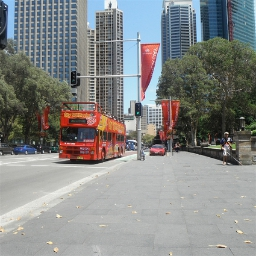}} &
         {\includegraphics[width=0.18\textwidth]{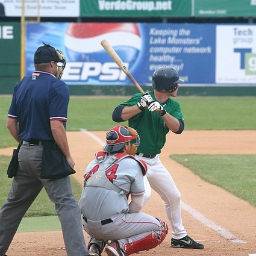}} \\
         (a) toilet & (b) tower & (c) giraffe & (d) street & (e) player \\
    \end{tabular}
    \caption{\textbf{Example NSD Images.} Below, we present detailed captions generated by MLLMs using images as input, alongside captions from our method using different feature spaces with brain signals as input.}
    \label{fig:supmat_nsd_test}
\end{figure}

We provide COCO~\cite{lin2014microsoft} captions and detailed captions for example images (as shown in~\cref{fig:supmat_nsd_test}) from the NSD test set. Results from LLaVA-v1.5 7B (LLaVA)~\cite{liu2023visual} serve as pseudo ground truth for detailed captioning evaluation, while others, including LLaVA-MoF (MoF)~\cite{tong2024eyes}, DenseConnector-v1.5 7B (DC)~\cite{yao2025dense}, and Matryoshka-MM-v1.5 7B (M3)~\cite{cai2024matryoshka} are for reference.

\vspace{8pt}
\begin{itemize}[leftmargin=17.5mm]
\setlength{\itemsep}{2pt}
    \item[\textbf{MLLM}] \textbf{Caption}
    \vspace{3pt}
    \small{
    \item[COCO] (a) (14 words) A very small bathroom stall with a toilet and several rolls of toilet paper.
    \item[]     (b)  (10 words) A very tall clock tower with two clocks on it.
    \item[]     (c) (10 words) A large giraffe walking across a dirt lot near trees.
    \item[]     (d) (12 words) A red and yellow double decker bus on street next to trees.
    \item[]     (e) (10 words) A baseball player holding a bat next to home plate.
    \item[LLaVA] (a)  (44 words) The image shows a small bathroom with a white toilet sitting on a tiled floor. The toilet is positioned in the corner of the room, and there is a toilet paper roll on the wall nearby. The bathroom appears to be clean and well-maintained.
    \item[]     (b)  (59 words) The image features a tall clock tower with two clocks on its side. The clocks are positioned at different heights, one higher than the other. The tower is made of stone and has a prominent presence in the scene. The clocks are visible from different angles, making it easy to see the time from various points around the tower.
    \item[]     (c) (60 words) The image features a giraffe walking in a dirt area, possibly a zoo enclosure. The giraffe is the main focus of the scene, with its long neck and legs prominently visible. The enclosure is surrounded by a fence, and there is a building nearby. The giraffe appears to be walking towards the building, possibly seeking shelter or exploring its surroundings.
    \item[]     (d) (114 words)  The image features a red double-decker bus driving down a city street, surrounded by tall buildings. The bus is positioned in the middle of the scene, and there are several cars on the road as well. Some cars are behind the bus, while others are in front of it.
    
    There are multiple traffic lights along the street, with one near the left side of the bus, another further down the road, and a third one on the right side. A few pedestrians can be seen walking along the sidewalk, with one person closer to the bus and two others further away. Additionally, there is a handbag visible near the right side of the scene.

    \item[]     (e) (114 words) The image captures a baseball game in progress, with a batter holding a baseball bat and preparing to swing. The batter is wearing a green shirt and is positioned near the center of the scene. The catcher, wearing a red shirt, is crouching behind the batter, ready to catch the ball. 
    
    There are several other people in the scene, including a person standing to the left of the batter and another person on the right side. A baseball glove can be seen in the hands of one of the players, and a sports ball is visible in the air. The scene is set on a baseball field, with a bench located in the background.

    \item[MoF] (a) (67 words)   The image features a small, clean bathroom with a white toilet situated in the corner. The toilet is equipped with a control panel and a remote, which can be used to operate the toilet. There are also several rolls of toilet paper placed on the wall above the toilet, ensuring that users have easy access to them. The bathroom appears to be well-maintained and ready for use.

    \item[] (b) (58 words)  The image features a tall clock tower with two clocks on its sides. The clocks are positioned at different heights, with one being higher up and the other lower down. The tower is made of stone and has a brown color. The clocks are visible from various angles, making it easy to tell the time from different directions.

    \item[] (c) (60 words)  The image features a giraffe standing in a dirt area, possibly a zoo enclosure, with a blue and white striped wall in the background. The giraffe is walking towards the camera, and its long neck and legs are prominently visible. The giraffe appears to be the main focus of the scene, with no other animals or people in the image.

    \item[] (d) (102 words)  The image features a large red double-decker bus driving down a city street, surrounded by tall buildings. The bus is positioned in the middle of the scene, with its height extending from the lower to the upper part of the image. 
    
    There are several people walking along the sidewalk, with some closer to the bus and others further away. A few cars can be seen on the street, with one car located behind the bus and another one in front of it. A traffic light is visible on the left side of the image, indicating the flow of traffic in the city.

    \item[] (e) (78 words)  The image captures a baseball game in progress, with a batter standing at home plate, holding a baseball bat and preparing to swing. The catcher and umpire are also present, with the catcher wearing a baseball glove. 
    
    There are several other people in the scene, including teammates and possibly coaches, scattered around the field. A baseball bat can be seen lying on the ground, and a sports ball, likely the baseball, is visible in the scene as well.
    
    \item[DC] (a) (64 words)  The image shows a small bathroom with a white toilet positioned in the middle of the room. The toilet is equipped with a handicap rail, making it accessible for people with disabilities. There is a toilet paper roll on the wall next to the toilet, and a towel is placed on the back of the toilet. The bathroom appears to be clean and well-maintained.

    \item[] (b) (77 words) The image features a tall clock tower with two clocks on its side, one on the left and the other on the right. The clocks are positioned at different heights, with the one on the left being higher than the one on the right. The tower is part of a large building, and the clocks are visible from various angles. The sky in the background is clear and blue, providing a beautiful contrast to the clock tower.

    \item[] (c) (52 words) The image features a giraffe walking in a dirt area near a building. The giraffe is the main focus of the scene, with its long neck and legs visible as it moves across the ground. The building appears to be a part of the giraffe's enclosure, providing a backdrop for the scene.

    \item[] (d) (106 words) The image depicts a city street with a red double-decker bus driving down the road. The bus is surrounded by tall buildings, giving the impression of a busy urban environment. There are several cars on the street, including one behind the bus and others parked or driving in the same direction.
    
    There are multiple traffic lights visible in the scene, with one near the left side of the street and others scattered throughout the area. A few pedestrians can be seen walking along the sidewalk, with one person carrying a handbag. The overall atmosphere of the image is bustling and lively, typical of a city street.

    \item[] (e) (102 words) The image captures a baseball game in progress, with a batter holding a baseball bat and preparing to swing. The catcher and the umpire are also present, closely observing the batter's actions. The batter is wearing a green shirt and is positioned near the center of the scene. The catcher is located to the left of the batter, while the umpire stands to the left of the catcher.
    
    In the background, there are two more people, possibly teammates or spectators, standing further away from the action. The scene is set on a baseball field, with a baseball glove visible in the foreground.

    \item[M3] (a) (34 words) The image shows a small bathroom with a white toilet. The toilet is situated in the corner of the room, and there is a sink nearby. The bathroom appears to be clean and well-maintained.

    \item[] (b) (53 words)  The image features a large clock tower with two clocks on its side. The clocks are visible on different sides of the tower, making it a prominent and functional structure. The tower is part of a building, and the clocks are placed at various heights, adding to the tower's overall height and prominence.

    \item[] (c) (44 words) A giraffe is walking in a zoo enclosure, surrounded by a fence. It is standing near a building and appears to be looking at the camera. The giraffe is the main focus of the image, and it occupies a significant portion of the scene.

    \item[] (d) (82 words) A city street with a red double-decker bus driving down the road. The bus is passing by a traffic light and a pedestrian crossing. There are several cars on the street, including one behind the bus and another one further down the road. A few people can be seen walking or standing on the sidewalk, with one person carrying a handbag. The scene captures the hustle and bustle of city life, with various modes of transportation and pedestrians going about their day.

    \item[] (e) (96 words) In the image, a baseball player is at home plate, holding a baseball bat and preparing to swing. The catcher is positioned behind the batter, wearing a baseball glove, ready to catch the ball. The umpire is also present, closely observing the game.
    
    There are several other people in the scene, some of whom are standing near the edge of the field, possibly waiting for their turn to play or watching the game. A bench can be seen in the background, likely for players to sit on when they are not actively participating in the game.
    }
\end{itemize}
\vspace{2mm}

\subsection{Salient QA Examples}
\label{subsec:supmat_example_qa}

This section presents constructed Salient QA examples using images in~\cref{fig:supmat_nsd_test} as references.

\vspace{8pt}
\begin{itemize}[leftmargin=17.5mm]
\setlength{\itemsep}{2pt}
    \item[\textbf{Reference}] \textbf{Salient QA}
    \vspace{3pt}
    \small{
    \item[\cref{fig:supmat_nsd_test}(a)] Q: Which description best fits the `bathroom' in the image? A. The bathroom is narrow. B. The bathroom is huge. C. The bathroom is dirty. A: ``A''.
    \item[\cref{fig:supmat_nsd_test}(b)] Q: How is the weather? A. Cloudy. B. Sunny. C. Rainy. A: ``B''.
    \item[\cref{fig:supmat_nsd_test}(c)] Q: Where is the animal? A: On the street. B. In a zoo. C. In the forest. A: ``B''.
    \item[\cref{fig:supmat_nsd_test}(d)] Q: Which description best fits the `bus' in the image? A. The bus is blue. B. The bus is double decker. C. The bus is green. A: ``B''.
    \item[\cref{fig:supmat_nsd_test}(e)] Q: Which description best fits the `batter' in the image? A. He is wearing a black shirt. B. He is wearing shorts. C. He is wearing a hat. A: ``C''.
}
\end{itemize}
    
\section{Additional Experiments}
\label{sec:supmat_more_results}

This section provides more results on concise captioning comparison (\cref{subsec:supmat_concise_caption}), detailed captioning comparison (\cref{subsec:supmat_detailed_caption}), visual reconstruction (\cref{subsec:supmat_reconstruction}), denoiser training visualization (\cref{subsec:supmat_denoiser_as_tabilizer}), and UMAP visualization (\cref{subsec:supmat_umap}).

\subsection{Concise Captioning Comparison}
\label{subsec:supmat_concise_caption}

\cref{tab:supmat_concise_caption} provides the COCO captions alongside the concise captions predicted by our method, VINDEX (built upon Shikra~\cite{chen2023shikra}), as well as state-of-the-art baselines: SDRecon~\cite{takagi2023improving}, BrainCap~\cite{ferrante2023brain}, OneLLM~\cite{han2024onellm}, MindEye2~\cite{scotti2024mindeye2}, and UMBRAE~\cite{xia2024umbrae}.

\begin{table*}[thbp]
\centering
\vspace{-2mm}
\caption{\textbf{Concise Captioning Comparison.}
Each image is shown with concise captions from SDRecon~\cite{takagi2023improving}, BrainCap~\cite{ferrante2023brain}, OneLLM~\cite{han2024onellm}, MindEye2~\cite{scotti2024mindeye2}, MEVOX~\cite{xia2025mevox}, UMBRAE~\cite{xia2024umbrae}, and VINDEX (built upon Shikra~\cite{chen2023shikra}). Refer to~\cref{subsec:supmat_example_caption} for captions from COCO.
}
\label{tab:supmat_concise_caption}
{
\begin{tabular}{cc}
\toprule
Image & Caption\\
\midrule
\makecell[c]{
\begin{minipage}[b]{0.2\linewidth}
    \centering
    {\includegraphics[width=3.5cm]{images/supmat/nsd_test/idx_19.png}}
\end{minipage}
} & \makecell[c]{
\begin{minipage}[b]{0.75\linewidth}
    {\small
    SDRecon: a small room in the white bathroom room fitted bathroom, hall room fitted modern bathroom
    \vspace{0.45mm}\\
    BrainCap: a bathroom with a sink and a toilet
    \vspace{0.45mm}\\
    OneLLM: A bathroom with a white toilet and a white sink.
    \vspace{0.45mm}\\
    MindEye2: a bathroom with a toilet and a sink.
    \vspace{0.45mm}\\
    MEVOX: A bathroom with a toilet and a sink.
    \vspace{0.45mm}\\
    UMBRAE-S: A bathroom with a toilet, sink and mirror.
    \vspace{0.45mm}\\
    UMBRAE: A bathroom with a toilet, sink and shower.
    \vspace{0.45mm}\\
    VINDEX-S: A bathroom with a toilet, a shower and a tub.
    \vspace{0.45mm}\\
    VINDEX: A white bathroom with a toilet and a shower.
    }
\end{minipage}
} \\
\midrule
\makecell[c]{
\begin{minipage}[b]{0.2\linewidth}
    \centering
    {\includegraphics[width=3.5cm]{images/supmat/nsd_test/idx_22.png}}
\end{minipage}
} & \makecell[c]{
\begin{minipage}[b]{0.75\linewidth}
    {\small
    SDRecon: a bathroom 
    \vspace{0.45mm}\\
    BrainCap:  a clock on the side of a tower.
    \vspace{0.45mm}\\
    OneLLM: A large clock tower sitting on top of a building.
    \vspace{0.45mm}\\
    MindEye2: a clock tower with a tower in the background.
    \vspace{0.45mm}\\
    MEVOX: A tall building with a clock on the top.
    \vspace{0.45mm}\\
    UMBRAE-S: An old building with a clock on the top.
    \vspace{0.45mm}\\
    UMBRAE: A clock tower that has two clocks sits in the sky.
    \vspace{0.45mm}\\
    VINDEX-S: A clock tower on a building with a steeple on top.
    \vspace{0.45mm}\\
    VINDEX: A large building with a clock on the top.
    }
\end{minipage}
} \\
\midrule
\makecell[c]{
\begin{minipage}[b]{0.2\linewidth}
    \centering
    {\includegraphics[width=3.5cm]{images/supmat/nsd_test/idx_41.png}}
\end{minipage}
} & \makecell[c]{
\begin{minipage}[b]{0.75\linewidth}
    {\small
    SDRecon: a wild park in the woods with two cars parked
    \vspace{0.45mm}\\
    BrainCap: a group of trees and a zebra
    \vspace{0.45mm}\\
    OneLLM: a fire truck parked in front of a building.
    \vspace{0.45mm}\\
    MindEye2: a giraffe standing in a field.
    \vspace{0.45mm}\\
    MEVOX: A zebra standing in the middle of a lush green field.
    \vspace{0.45mm}\\
    UMBRAE-S: A giraffe is standing in a grassy field.
    \vspace{0.45mm}\\
    UMBRAE: A giraffe is standing in a grassy field.
    \vspace{0.45mm}\\
    VINDEX-S: Two giraffes standing next to each other in the grass.
    \vspace{0.45mm}\\
    VINDEX: A giraffe standing in a grassy field next to trees.
    }
\end{minipage}
} \\
\midrule
\makecell[c]{
\begin{minipage}[b]{0.2\linewidth}
    \centering
    {\includegraphics[width=3.5cm]{images/supmat/nsd_test/idx_47.png}}
\end{minipage}
} & \makecell[c]{
\begin{minipage}[b]{0.75\linewidth}
    {\small
    SDRecon: a street in downtown, the building is located at the center of this street
    \vspace{0.45mm}\\
    BrainCap: a city street with a lot of cars and a bus. 
    \vspace{0.45mm}\\
    OneLLM: A long red train car sits on train tracks.
    \vspace{0.45mm}\\
    MindEye2: a street with a car and a bus.
    \vspace{0.45mm}\\
    MEVOX: A bus driving down a street next to a tall building.
    \vspace{0.45mm}\\
    UMBRAE-S: A street with cars parked on the side of it.
    \vspace{0.45mm}\\
    UMBRAE: A city street filled with lots of traffic.
    \vspace{0.45mm}\\
    VINDEX-S: A street view of cars driving down the street.
    \vspace{0.45mm}\\
    VINDEX: A city street filled with lots of traffic.
    }
\end{minipage}
} \\
\midrule
\makecell[c]{
\begin{minipage}[b]{0.2\linewidth}
    \centering
    {\includegraphics[width=3.5cm]{images/supmat/nsd_test/idx_57.png}}
\end{minipage}
} & \makecell[c]{
\begin{minipage}[b]{0.75\linewidth}
    {\small
    SDRecon: two men on the field, one is wearing a baseball bat and the other player
    \vspace{0.45mm}\\
    BrainCap: a young man holding a baseball bat.
    \vspace{0.45mm}\\
    OneLLM: A group of young men standing on a tennis court.
    \vspace{0.45mm}\\
    MindEye2: a baseball player is standing in front of a batter.
    \vspace{0.45mm}\\
    MEVOX: A man holding a tennis racket on a tennis court.
    \vspace{0.45mm}\\
    UMBRAE-S: A baseball player getting ready to bat.
    \vspace{0.45mm}\\
    UMBRAE: A group of men playing baseball on a field.
    \vspace{0.45mm}\\
    VINDEX-S: A baseball player is getting ready to hit the ball.
    \vspace{0.45mm}\\
    VINDEX: A baseball player getting ready to swing at the ball.
    }
\end{minipage}
} \\
\bottomrule
\end{tabular}
}
\end{table*}

\subsection{Detailed Captioning Comparison}
\label{subsec:supmat_detailed_caption}

Below is a comparison of detailed captioning results using the four feature spaces—SE, ME, AF, and NF.
Example images from the NSD test set are shown in~\cref{fig:supmat_nsd_test}. AF typically generates nonsensical outputs, such as garbled text and blank spaces, and therefore the results are not presented here.

\vspace{8pt}
\begin{itemize}[leftmargin=17.5mm]
\setlength{\itemsep}{2pt}
    \item[\textbf{Feature}] \textbf{Detailed Captioning}
    \vspace{3pt}
    \small{
    \item[SE] (a) (57 words) The image features a bathroom with a white toilet situated in the middle of the room. The toilet is surrounded by a tiled floor, which gives the bathroom a clean and modern appearance. The walls of the bathroom are also tiled, adding to the overall cohesive design. The bathroom appears to be well-maintained and ready for use.
    \item[] (b) (79 words) The image features a tall, old building with a clock tower on top. The clock is prominently displayed on the side of the building, making it a focal point. The building has a steeple-like structure, giving it a unique and historic appearance. The clock is positioned towards the center of the building, drawing attention to its presence. The overall architecture of the building suggests that it may be a church or a similar type of structure with historical significance.

    \item[] (c) (75 words) The image features a large giraffe standing in a grassy field. The giraffe is the main focus of the scene, and it appears to be walking through the field. The field is lush and green, providing a natural habitat for the giraffe. The giraffe's long neck and legs are clearly visible, showcasing its unique features. The overall atmosphere of the image is serene and peaceful, capturing the beauty of the giraffe in its natural environment.

    \item[] (d) (115 words) The image depicts a busy city street scene with several cars and buses driving down the road. There are multiple cars in various positions, some closer to the foreground and others further back. A bus is also visible in the middle of the scene, adding to the traffic.
    
    In addition to the vehicles, there are several people walking along the sidewalk, going about their daily activities. Some of them are closer to the foreground, while others are further back in the scene.
    
    The street is lined with trees, providing a touch of greenery to the urban environment. The combination of the bustling traffic and the presence of pedestrians creates a lively atmosphere in the city.
    
    \item[] (e) (56 words) The image depicts a man wearing a baseball uniform, standing on a field with a baseball glove on his hand. He appears to be a baseball player, possibly waiting for a pitch or preparing to catch a ball. The scene takes place on a baseball field, with the man being the main focus of the image.

    \item[ME] (a) (40 words) The image features a white bathroom with a toilet and a sink. The toilet is located on the left side of the bathroom, while the sink is situated on the right side. The bathroom appears to be clean and well-maintained.

    \item[] (b) (86 words) The image features a large building with a clock tower, which is situated in the middle of a city. The clock tower is visible on the left side of the building, and the building itself is quite tall. The scene is set against a backdrop of trees, creating a picturesque view. The trees are scattered throughout the scene, with some located near the building and others further away. The combination of the clock tower, the building, and the trees creates a visually appealing and urban landscape.

    \item[] (c) (52 words) The image features a herd of zebras grazing in a grassy field. There are at least 13 zebras visible in the scene, scattered throughout the field. Some zebras are closer to the foreground, while others are further in the background. The zebras are peacefully eating grass, creating a serene and natural atmosphere.

    \item[] (d) (54 words) The image shows a city street with several cars parked along the side of the road. The cars are of various sizes and are parked in a row. The street appears to be empty, with no people visible in the scene. The cars are parked in a line, creating an organized and orderly appearance.

    \item[] (e) (69 words) The image features a tennis court with a tennis player in action. The player is holding a tennis racket and is in the middle of a swing, likely returning a volley. The tennis player is positioned towards the left side of the court. The court is surrounded by a fence, and there are several people in the background, possibly watching the game or waiting for their turn to play.
    
    \item[NF] (a) (97 words) The image features a bathroom with a white toilet situated in the corner of the room. The toilet is positioned under a window, which allows natural light to enter the space. The bathroom also has a sink, which is located towards the right side of the room.
    
    There are two bottles in the bathroom, one placed near the sink and the other closer to the toilet. Additionally, there is a cup on the left side of the room, and a bowl can be seen near the sink. The overall appearance of the bathroom is clean and well-organized.

    \item[] (b) (79 words) The image features a large, old-fashioned clock tower with a steeple, situated on top of a building. The clock is prominently displayed on the tower, making it a focal point of the scene. The tower is surrounded by a group of trees, creating a picturesque setting.
    
    There are several people in the scene, with some standing closer to the clock tower and others further away. They appear to be enjoying the view of the tower and the surrounding environment.

    \item[] (c) (46 words) The image features a large giraffe standing in a grassy field, surrounded by trees. The giraffe appears to be walking through the grass, possibly in search of food. The scene is set in a natural environment, with the giraffe being the main focus of the image.

    \item[] (d) (104 words) The image depicts a city street with a white car parked on the side of the road. The car is positioned near the center of the scene, and it appears to be a compact vehicle. There are several other cars parked along the street, with some closer to the foreground and others further in the background.
    
    In addition to the cars, there are two people visible in the scene. One person is standing near the left side of the image, while the other person is located closer to the center. The street is lined with trees, providing a pleasant atmosphere for the city setting.

    \item[] (e) (101 words) The image features a baseball field with several baseball players standing on the field. There are at least nine people visible in the scene, with some of them closer to the foreground and others further in the background. A baseball glove can be seen on the ground, indicating that the players are either preparing for a game or have just finished one.
    
    The players are spread out across the field, with some standing closer to the center and others near the edges. The overall atmosphere of the scene suggests that the players are engaged in a casual or recreational baseball game.
}
\end{itemize}

\subsection{Visual Reconstruction}
\label{subsec:supmat_reconstruction}

This paper explores fMRI-based multimodal interaction using MLLMs, focusing on perception over reconstruction. We address four tasks: Concise Captioning, Descriptive Captioning, and Concept Localization, and Complex Reasoning. Despite this focus, the generated multimodal explanations~\cite{shen2024neuro,xia2024umbrae} are shown to improve reconstruction performance in \cref{tab:supmat_vis_recon}.

\begin{table}[htbp]
  \centering
  \caption{\textbf{Quantitative Visual Reconstruction Evaluation.} We report metrics following prior studies~\cite{ozcelik2023brain,scotti2023reconstructing}.
  }
  \resizebox{\linewidth}{!}{
    \begin{tabular}{l|c|cccc|cccc}
      \toprule
      \multirow{2}{*}{Method} & \multirow{2}{*}{\#Models}   & \multicolumn{4}{c|}{Low-Level}   & \multicolumn{4}{c}{High-Level}   \\
      ~ & ~ & PixCorr $\uparrow$     & SSIM $\uparrow$      & AlexNet(2) $\uparrow$ & AlexNet(5) $\uparrow$         & Inception $\uparrow$        & CLIP $\uparrow$   & EffNet-B $\downarrow$     & SwAV $\downarrow$         \\
      \midrule
      Mind-Reader~\cite{lin2022mind}    & 4 & -    & -  & -      & - & 78.2\%      & -  & -& -\\
      SDRecon~\cite{takagi2023improving}      &  4  & -    & -  & 83.0\%& 83.0\%        & 76.0\%      & 77.0\%      & -& -\\
      Brain-Diffuser~\cite{ozcelik2023brain}    & 4 & .254 & \underline{.356}   & 94.2\%     & 96.2\%         & 87.2\%       & 91.5\%       & .775       & .423       \\
      MindEye~\cite{scotti2023reconstructing}     & 4 & \textbf{.309 }      & .323  & \underline{94.7\%}     & \textbf{97.8\%}  & 93.8\% & 94.1\%      & .645 & \underline{.367}  \\
      DREAM~\cite{xia2024dream}      & 4 & \underline{.288}     & .338   & \textbf{95.0\%}       & \underline{97.5\%} & \underline{94.8\%} & \underline{95.2\%}  & \underline{.638} & .413 \\
      \midrule
      MindBridge~\cite{wang2024mindbridge}        & 1 & .151 & .263 & 87.7\%  & 95.5\%         & 92.4\%       & 94.7\%       & .712       & .418       \\
      UMBRAE~\cite{xia2024umbrae}    & 1 & .283     & .328  & 93.9\% & 96.7\%         & 93.4\%       & 94.1\% & .700       & .393       \\
      NeuroVLA~\cite{shen2024neuro}        & 1 & .265  & \textbf{.357}      & 93.1\%  & 97.1\% & \textbf{96.8\%} & \textbf{97.5\%} & \textbf{.633} & \textbf{.321} \\
      VINDEX & 1 & .203 & .317 & 93.5\% & 96.9\% & 93.5\% & 95.1\% &  .658   & .403 \\
      \midrule
    \end{tabular}
    \label{tab:supmat_vis_recon}
  }
\end{table}

\subsection{Denoiser as a Training Stabilizer} %
\label{subsec:supmat_denoiser_as_tabilizer}

\begin{figure}[thbp]
\centering
\renewcommand{\arraystretch}{0.5}
\setkeys{Gin}{width=0.22\linewidth}
\setlength{\tabcolsep}{1.2pt}
{
\begin{tabular}{cc}
\includegraphics[width=0.45\textwidth]{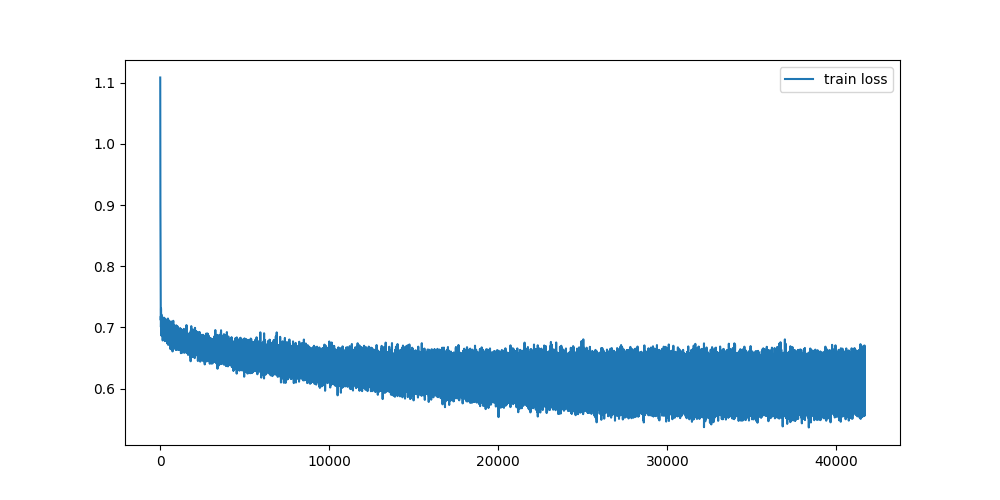} & \includegraphics[width=0.45\textwidth]{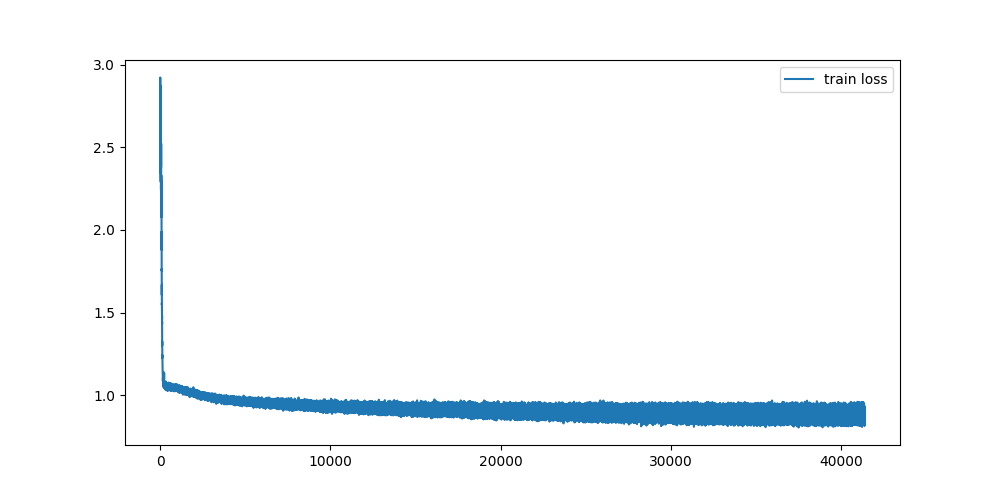} \\
(a) & (b)  \\
\end{tabular}
\caption{\textbf{Denoiser as a Training Stabilizer.} (a) The vanilla regression loss decreases but exhibits significant oscillation; (b) The training process becomes less oscillatory with the incorporation of diffusion loss, which stabilizes the training and accelerates convergence.}
\label{fig:supmat_denoiser_as_tabilizer}
}
\end{figure}

We conduct an ablation study on the denoiser architecture and weights in the main paper, working together with regression loss to improve performance. Here, we provide further analysis on how the denoiser stabilizes training. As shown in~\cref{fig:supmat_denoiser_as_tabilizer}, the vanilla regression loss decreases but exhibits significant oscillation. Incorporating diffusion loss stabilizes the training process and leads to faster convergence.

\subsection{UMAP Visualization}
\label{subsec:supmat_umap}

To better understand the brain-feature alignment, we apply UMAP~\cite{mcinnes2018umap} for dimensionality reduction, projecting the predicted features from brain encoders and target vision encoders into a 2D space. The well-aligned features should form cohesive clusters, while misaligned features are expected to be disjointed~\cite{scotti2023reconstructing}.

\begin{figure}[thbp]
\centering
\renewcommand{\arraystretch}{0.5}
\setkeys{Gin}{width=0.22\linewidth}
\setlength{\tabcolsep}{1.2pt}
{
\begin{tabular}{cccc}
\includegraphics[width=0.25\textwidth]{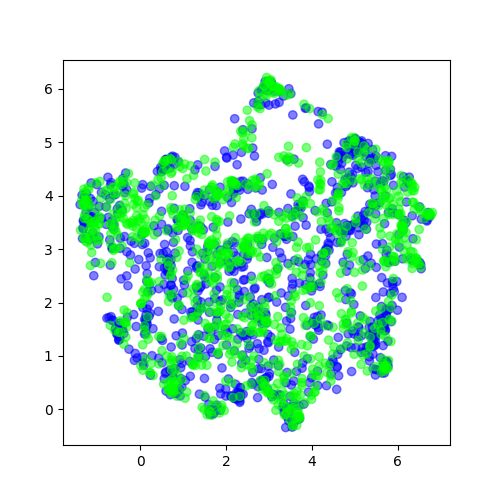} & \includegraphics[width=0.25\textwidth]{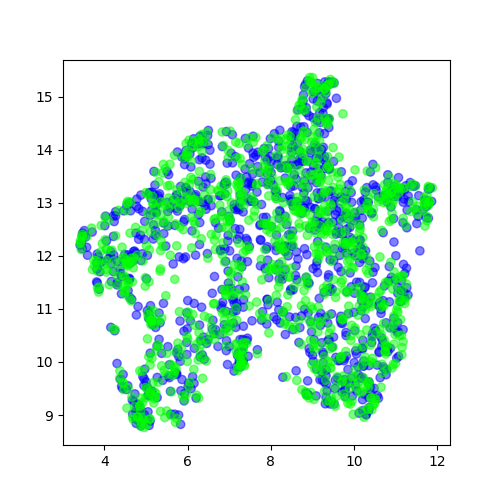} &
\includegraphics[width=0.25\textwidth]{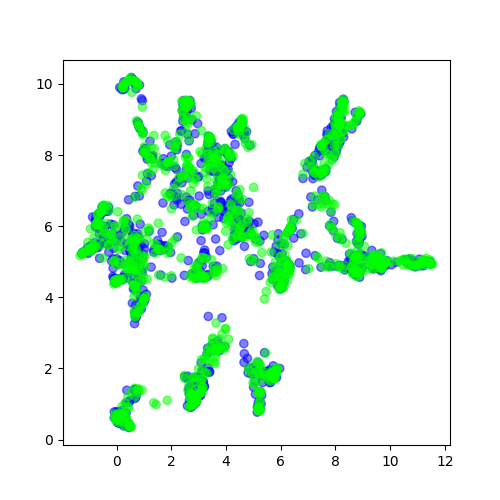} & \includegraphics[width=0.25\textwidth]{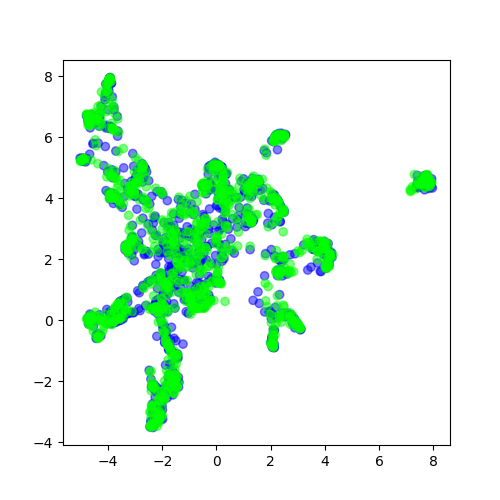} \\
(a) \textsc{b-clip224} & (b) \textsc{b-clip336} & (c) \textsc{b-dino224} & (d)\textsc{b-siglip384} \\
\end{tabular}
\caption{\textbf{UMAP Comparison.} The UMAP visualizations depict predicted features (blue) from brain encoders \textsc{b-clip224}, \textsc{b-clip336}, \textsc{b-dino224}, and \textsc{b-siglip384} using brain inputs, along with the corresponding image features (green) extracted from target vision encoders using associated images as input. For further details on the brain encoders, please refer to \cref{tab:pre-trained_brain_encoder}.}
\label{fig:supmat_umap}
}
\end{figure}

\cref{fig:supmat_umap} shows the UMAP visualization for predicted features from brain encoders \textsc{b-clip224}, \textsc{b-clip336}, \textsc{b-dino224}, and \textsc{b-siglip384} (fMRI as input), along with the corresponding ground truth features from target vision encoders (associated visual stimuli as input). Refer to \cref{tab:pre-trained_brain_encoder} for more details on pre-trained brain encoders.

\section{Discussion}
\label{sec:supmat_discussion}

Our method explores fMRI-based multimodal interaction using MLLMs but has certain limitations. First, we primarily focus on representative feature spaces  to validate the essence of our idea rather than exhaustively experimenting with all existing vision encoders. Second, we rely solely on fMRI data~\cite{allen2022massive} without extending our analysis to other neuroimaging techniques such as EEG or MEG. EEG, being more portable and cost-effective, enables large-scale data collection, albeit with lower spatial resolution. Incorporating EEG experiments could further explore alignment with image features while benefiting from larger data scales. Third, we align brain signals with image features, serving as a simple approach for zero-shot learning. However, as shown in~\cref{tab:pre-trained_brain_encoder} and~\cref{tab:supmat_vision_expert}, token numbers and feature dimensions increase when using vision encoders with higher image resolutions, imposing significant computational costs during training. Furthermore, results in the main paper indicate that aligning brain signals with higher-resolution visual features does not necessarily yield better understanding or reconstruction. Instead, aligning with tokens directly offers a more direct manner, enabling the use of token pruning and merging techniques~\cite{bolya2022token,kim2024token} to dynamically reduce token count based on the task, thereby alleviating the computational burden.

\paragraph{Visual Encoders}

Beyond the vision experts used in the main paper, there are several other pre-trained visual encoders that can be incorporated into the toolbox. These encoders, trained on various tasks and resolutions, allow us to explore the distinct advantages of different experts. We compile a set of vision experts, including:
(1) Constrastive Vision-Language Alignment: CLIP \cite{radford2021learning} and ConvNeXt~\cite{woo2023convnext} from OpenCLIP \cite{schuhmann2022laionb}.
(2) Visual Grounding: DINOv2~\cite{oquab2023dinov2} using self-supervised learning.
(3) Object-Centric Training: EVA-02~\cite{fang2023eva02} and CoDETR~\cite{zong2023detrs}, pre-trained on detection datasets.
(4) Optical Character Recognition (OCR): Pix2Struct \cite{lee2023pix2struct}.
(5) Segmentation: SAM~\cite{kirillov2023segment}. 
(6) Video-Language Pretraining: LanguageBind~\cite{zhu2023languagebind}.
(7) Vision Foundation Model: InternViT~\cite{chen2024internvl}.
The detailed task (taxonomy), input image size (resolution), and checkpoint for each vision encoder are in~\cref{tab:supmat_vision_expert}.
Preliminary results from recent hybrid multimodal models~\cite{shi2024eagle,chung2025unifying,zong2024mova,azadani2025leo} indicate that MLLMs using these task-specific vision encoders achieve optimal performance within their pre-training domains. For example, EVA-02~\cite{fang2023eva02} excels in the the visual question answering benchmark GQA~\cite{hudson2019gqa} and the object hallucination evaluation benchmark POPE~\cite{li2023evaluating}. Both CLIP~\cite{radford2021learning} and ConvNeXt~\cite{woo2023convnext} perform well across several benchmarks, benefiting from large-scale image-text pair training using contrastive loss. In contrast, while Pix2Struct excels at text recognition, it shows limited capability in object recognition and general VQA tasks. DINOv2~\cite{oquab2023dinov2} and SAM~\cite{kirillov2023segment}, trained via self-supervised learning and semantic segmentation, respectively, struggle with text recognition tasks.

\begin{table}[t!]
    \centering
    \caption{\textbf{Detailed on pre-trained Vision Experts.} These vision models are trained and specialized for specific tasks, and it has been shown that MLLMs using these task-specific vision encoders achieve optimal performance within their pre-training domains.}
    \label{tab:supmat_vision_expert}
    \begin{tabular}{llccl}
    \toprule
    Vision Expert & Task & Size & \#Dimension & Link\\ 
    \midrule
    CLIP~\cite{radford2021learning} & Image-text Contrastive & 448 & 1024 & \href{https://huggingface.co/openai/clip-vit-large-patch14-336}{openai/clip-vit-large-patch14-336}\\
    DINOv2~\cite{oquab2023dinov2}  & Visual Grounding &  448 & 1024 &  \href{https://github.com/facebookresearch/dinov2/blob/main/MODEL_CARD.md}{facebookresearch/dinov2}\\
    SAM~\cite{kirillov2023segment}  & Image Segmentation &  1024 & 1024 & \href{https://huggingface.co/facebook/sam-vit-large}{facebook/sam-vit-large}\\
    EVA-02~\cite{fang2023eva02}  & Object Detection &  1024 & 1024 & \href{https://huggingface.co/Yuxin-CV/EVA-02/blob/main/eva02/det/eva02_L_coco_det_sys_o365.pth}{EVA-02-Large}\\
    ConvNeXt~\cite{woo2023convnext} & Image Classification & 1024 & 1024 & \href{https://huggingface.co/laion/CLIP-convnext_xxlarge-laion2B-s34B-b82K-augreg-soup}{laion/CLIP-convnext-xxlarge}\\
    \bottomrule
    \end{tabular}
\end{table}

\paragraph{Hybrid Multimodal Models.} 

Besides the feature spaces discussed in the main paper, there are also other hybrid MLLMs, especially those based on a mixture of vision experts. MLLMs that utilize these task-specific vision encoders deliver optimal performance within their respective pre-training domains.
We provide a reference list in~\cref{tab:supmat_hybrid_mllm} for these hybrid MLLMs.

From the brain results presented in the paper, we observe that aligning with multiple vision encoders does not only bring performance gains but also increases training complexity. Brain signals struggle with the precise location of concepts, especially for small, inconspicuous objects~\cite{xia2024umbrae}. It is predictable that aligning with SAM features~\cite{kirillov2023segment} may not effectively support decoding brain signals into clear segmentation results. These findings from both research directions inspire us to further explore aligning brain signals with foundation vision models such as ConvNeXt~\cite{woo2023convnext} or InternViT~\cite{chen2024internvl}, which support universal perception, rather than aligning with multiple vision experts specialized in individual vision tasks.

\setlength{\tabcolsep}{4pt}
\setlength{\fboxrule}{0pt} 
\setlength{\fboxsep}{2pt}
\begin{table}[t!]
\centering
\caption{\textbf{Details on Hybrid MLLMs with Mixtured Vision Encoders}.}
\label{tab:supmat_hybrid_mllm}
\begin{tabular}{lll}
\toprule
\textbf{Method}        & \textbf{Encoders}    & \textbf{MLLM}           \\ \midrule
EAGLE~\cite{shi2024eagle}     & CLIP, ConvNeXt, EVA-02, Pix2Struct, DINOv2, SAM& LLaVA-1.5/Qwen2.5       \\ 
MERV~\cite{chung2025unifying}      & DINOv2, ViViT, SigLIP, LanguageBind            & LLaMA2/3   \\ 
MoVA~\cite{zong2024mova}      & CLIP, DINOv2, CoDETR, SAM, Pix2Struct, \etc          & Vicuna/Llama3/Yi        \\ 
MoME~\cite{shen2025mome}      & CLIP, DINOv2, Pix2Struct          & Vicuna-v1.5\\ 
LEO~\cite{azadani2025leo}       & InternViT, SAM       & InternVL   \\ \bottomrule
\end{tabular}
\end{table}

{
    \small
    \bibliographystyle{ieeenat_fullname}
    \bibliography{reference}
}

\end{document}